\documentclass[journal]{IEEEtran}
\usepackage{cite}

\ifCLASSINFOpdf
  \usepackage[pdftex]{graphicx}
\else
  \usepackage[dvips]{graphicx}
\fi
\usepackage{amsmath}
\usepackage{bm}
\usepackage{amssymb} 
\usepackage{amsthm}
\newtheorem{thm}{Theorem}
\newtheorem{lemma}{Lemma}

\newtheorem{corol}{Corollary}

\usepackage{rotating}

\DeclareMathOperator*{\argmax}{argmax}
\DeclareMathOperator*{\argmin}{argmin}

\newcommand{\defineq}{\overset{\text{def}}{=}} 

\usepackage{booktabs}
\usepackage{enumerate}
\usepackage{algorithmic}
\usepackage{algorithm}

\usepackage{tikz}
\usepackage{smartdiagram}
\usetikzlibrary{arrows,shapes}
\usetikzlibrary{positioning,decorations.pathreplacing,shapes}
\usepackage[english]{babel}
\usepackage{microtype}
\usetikzlibrary{shapes.geometric,shapes.arrows,decorations.pathmorphing}
\usetikzlibrary{matrix,chains,scopes,positioning,arrows,fit}
\usepackage{color}

\usepackage{array}

\usepackage{threeparttable}
\usepackage{multirow}

\ifCLASSOPTIONcompsoc
  \usepackage[caption=false,font=normalsize,labelfont=sf,textfont=sf]{subfig}
\else
  \usepackage[caption=false,font=footnotesize]{subfig}
\fi
\usepackage{url}

\begin{document}

\newcommand{\citet}[1]{\citeauthor{#1} \shortcite{#1}}
\newcommand{\citep}{\cite} 
\newcommand{\citealp}[1]{\citeauthor{#1} \citeyear{#1}}

\newcommand*{\cancer}{\text{cancer}}
\newcommand*{\testp}{\text{test}+}

\definecolor{mypurple}{RGB}{217,78,121}
\definecolor{myblue}{RGB}{121,145,217}
\definecolor{mygreen}{RGB}{214,229,89}%
\definecolor{myyellow}{RGB}{214,229,89}%
\definecolor{myorange}{RGB}{214,165,105}

\def\etal{\emph{et al.}}
\def\ie{i.e.,}
\def\eg{e.g.,}

\def\sign{\mathop{\rm{sgn}}}
\def\Err{\mathop{\rm{Err}}}
\def\diver{\mathop{\rm{div}}}
\def\margin{\mathop{\rm{margin}}}
\def\dist{\mathop{\rm{cred}}} 

\newcommand{\pdfull}{Ensemble Pruning based on Diversity}
\newcommand{\pdabbr}{\emph{EPBD}}
\newcommand{\fdfull}{Ensemble Pruning Framework Utilizing the Trade-off Between Accuracy and Diversity}
\newcommand{\fdabbr}{\emph{FTAD}}
\newcommand{\pdtitle}{When does Diversity Help Generalization in Classification Ensembles?}

\title{\pdtitle}
\author{Yijun~Bian %
        and~Huanhuan~Chen,~\IEEEmembership{Senior~Member,~IEEE}
\thanks{Manuscript received October 27, 2019; revised July 31, 2020; revised November 04, 2020; accepted January 13, 2021. %
\emph{(Corresponding author: Huanhuan Chen.)}}%
\thanks{Y. Bian and H. Chen are with the School of Computer Science and Technology, University of Science and Technology of China (USTC), Hefei, 230027 China (e-mails: yjbian@mail.ustc.edu.cn; hchen@ustc.edu.cn).}}

\markboth{IEEE Transactions on Cybernetics,~Vol.~, No.~, Month~Year}%
{Bian \MakeLowercase{\textit{et al.}}: \pdtitle}

\maketitle

\begin{abstract}
Ensembles, as a widely used and effective technique in the machine learning community, succeed within a key element---``diversity.'' 
The relationship between diversity and generalization, unfortunately, is not entirely understood and remains an open research issue.
To reveal the effect of diversity on the generalization of classification ensembles, we investigate three issues on diversity, i.e., the measurement of diversity, the relationship between the proposed diversity and the generalization error, and the utilization of this relationship for ensemble pruning. 
In the diversity measurement, we measure diversity by error decomposition inspired by regression ensembles, which decomposes the error of classification ensembles into accuracy and diversity.
Then we formulate the relationship between the measured diversity and ensemble performance through the theorem of margin and generalization and observe that the generalization error is reduced effectively only when the measured diversity is increased in a few specific ranges, while in other ranges larger diversity is less beneficial to increasing the generalization of an ensemble. 
Besides, we propose two pruning methods based on diversity management to utilize this relationship, which could increase diversity appropriately and shrink the size of the ensemble without much-decreasing performance. 
Empirical results validate the reasonableness of the proposed relationship between diversity and ensemble generalization error and the effectiveness of the proposed pruning methods. 
\end{abstract}
\begin{IEEEkeywords}
Ensemble learning, diversity, ensemble pruning, error decomposition.
\end{IEEEkeywords}

\IEEEpeerreviewmaketitle

\section{Introduction}

\IEEEPARstart{E}{nsemble} learning has attracted plenty of research attention in the machine learning community thanks to its remarkable potential \citep{zhou2012ensemble}, and it has been widely used in many real-world applications such as object detection, object recognition, object tracking \citep{girshick2014rich,wang2012mining,zhou2014ensemble,ykhlef2017efficient}, fault diagnosis, malware detection, depression detection \cite{06358882,07274666,08217789}, and so on \cite{6609055,6392473,6069610,7464340,8276629,8976239,8930086,8721696,6819051,9113329,9198926}. 
Rather than relying on one single model, an ensemble is a set of learned models that make decisions collectively. 
It is widely accepted that an ensemble usually generalizes better than one single model \citep{wu2008top,rokach2010ensemble,herrera2016ensemble,gu2015multi}.
Dietterich~\cite{dietterich2000ensemble} stated that an ensemble of classifiers succeeded in better accuracy if and only if its individual members are accurate and diverse. 
One classifier is accurate if its error rate is better than random guessing on new instances.
Two classifiers are diverse if they make different errors in new instances.
The diversity of an ensemble usually decreases with the increasing of accuracies of its members~\cite{lu2010ensemble}. 
Thus, how to handle the trade-off between accuracy and diversity appropriately is a crucial issue in ensemble learning.

Unfortunately, there is still no consensus in the community on the definition or measurement for diversity, unlike the precise accuracy. 
Existing ensemble methods create different classifiers implicitly or heuristically by manipulating input data, input features, or output targets. 
Existing diversity measures are various, yet none of them show superiority over each other~\cite{tang2006analysis,gu2015multi,Tsymbal2005Diversity}. 
They are generally divided into pairwise and non-pairwise diversity measures~\cite{kuncheva2003diversity,zhou2012ensemble}. 
Lots of previous work made an effort to seek the role of diversity in ensemble learning \cite{brown2009information,zhou2010multi,yu2011diversity,jiang2017generalized}. 
In regression ensembles, the diversity is defined based on the error decomposition~\cite{krogh1995neural}, in which the error of regression ensembles is split into the \label{key}accuracy term and diversity term. 
The classic Ambiguity Decomposition \citep{krogh1995neural} and Bias-Variance-Covariance decomposition \citep{geman1992neural} are two commonly used error decomposition schemes.
However, both of them are only suitable for regression tasks with the square loss. 

Moreover, the relationship between diversity and generalization in the literature remains an open question. 
Some researchers, such as Kuncheva and Whitaker~\cite{kuncheva2003diversity}, hold doubts about the usefulness of diversity measures in building classification ensembles in real-life pattern recognition problems. 
Conversely, some researchers hold the view as well that diversity among the members of a team of classifiers is a crucial issue in classifiers' combination~\citep{li2012diversity}. 
They are two inconsistent views, so, weirdly, both of them have some supporting experimental results in some cases. 
For instance, some empirical results~\citep{kuncheva2003diversity} showed less correlation between diversity and generalization error by varying the diversity in the ensemble. 
Meanwhile, some experiments~\citep{yu2011diversity} showed precisely the opposite results. 
Hence, to clarify this issue, whether or not diversity affects the performance of generalization becomes particularly essential. 

Inspired by error decomposition of regression ensembles, 
we propose a measure of diversity using error decomposition for classification ensembles with the $0/1$ error function firstly, where the error of classification ensembles is split into two terms: accuracy and diversity. 
Like regression ensembles, the diversity term measures the difference among members of classification ensembles. 
Secondly, we propose the relationship between the proposed diversity and generalization using the proposed diversity measure based on~\cite{herbrich2001pac,herbrich2002pac}. 
By taking bagging and AdaBoost as examples of ensemble methods, experiments are conducted to validate the proposed relationship by varying the diversity in bagging and AdaBoost.
Thirdly, we propose an ensemble pruning method named as ``\pdfull{} (\pdabbr),'' which utilizes the proposed relationship between the proposed diversity and generalization to improve the ensemble's generalization. 
To extend \pdabbr{} to use other diversity measures, we propose an ensemble pruning framework named as ``\fdfull{} (\fdabbr)'' as well, which utilizes the trade-off between accuracy and diversity. 
Moreover, empirical results are presented to validate the effectiveness of \pdabbr{} and \fdabbr{}. 
The contributions in this paper are three-fold: 
\begin{itemize}
    \item A diversity measure based on error decomposition is proposed to quantify the difference among classification ensemble members, for the benefit of conducting theoretical analyses later. 
    \item The relationship between the proposed diversity and the ensemble generalization error is investigated and analyzed theoretically, which demonstrates that diversity has different impacts on the ensemble generalization in different ranges. 
    \item Two ensemble pruning methods are proposed to select a subset of the original ensemble, while one utilizes the proposed theoretical relationship and the other as an extension of the former utilizes the trade-off between accuracy and diversity. 
\end{itemize}

The rest of this paper is organized as follows. 
The related work is summarized in Section~\ref{related}. 
And then, the core investigations are presented in Section~\ref{methodology}, answering three issues about diversity: 
(1) the measurement of diversity and its corresponding properties in Section~\ref{propose:measure} and \ref{propose:property};
(2) the relationship between the proposed diversity and generalization error of ensembles in Section~\ref{propose:relation};
(3) the utilization of this relationship about diversity in Section~\ref{propose:pruning} and \ref{prop:EPBD,complexity}. 
Finally, the empirical results are presented in Section~\ref{experiment}, followed by the conclusion in Section~\ref{conclusion}.

\section{Related Work}
\label{related}

Diversity is considered intuitively as the difference among individual members in an ensemble, with several alternative names such as dependence, orthogonality, or complementarity~\cite{kuncheva2003diversity, zhou2012ensemble}. 
In this section, we introduce existing methods to generate diverse individual classifiers in ensemble learning.
Then we describe existing methods to measure diversity in ensemble learning. 
Finally, we summarize some research about the crucial role diversity plays in ensemble learning.

\subsection{Generating Diverse Individual Classifiers for Ensembles}

Most ensemble methods attempt to generate diverse classifiers implicitly or heuristically by manipulating the input data or features, while few of them manipulate output targets~\cite{dietterich2000ensemble, yu2011diversity}.
For example, bagging~\cite{breiman1996bagging} and Boosting (including many variants)~\cite{freund1995desicion,freund1997decision, freund1996experiments} manipulate input data to promote diversity by choosing different subsets of the training data during training; 
Random forest~\cite{breiman2001random} manipulates subsets of input data or features to create diversity and gives competitive results;
Rotation forest~\cite{rodriguez2006rotation} applies principal component analysis (PCA) on each subset as an improved method.
Moreover, neural network (NN) ensembles also create diversity using different initial weights, different architectures of the networks, or different learning algorithms. 
However, the diversity generated by these methods is not precise and could not tell how much diversity in their generating process or whether diversity in ensembles should be increased or not for the specific ensemble.

\subsection{Diversity Measurements in Ensembles}

Diversity is usually measured based on the classification results of individual members in an ensemble \cite{chandra2004divace,li2012diversity,zhou2010multi,tang2006analysis,roli2001methods, gu2014generating}. 
Existing diversity measures are generally divided into pairwise and non-pairwise diversity measures \cite{kuncheva2003diversity,zhou2012ensemble}.
Based on the coincident errors between pairs of individual classifiers, pairwise diversity represents the behavior if both of them predict an instance identically or disagree with each other.
In this case, the overall diversity is the average of all possible pairs~\cite{gu2015multi}.
Pairwise diversity includes $Q$-statistic~\cite{yule1900vii}, $\kappa$-statistic~\cite{cohen1960coefficient}, disagreement measure~\cite{skalak1996sources,ho1998random}, correlation coefficient~\cite{sneath1973numerical}, double-fault~\cite{giacinto2001design}, and the hamming distance (HD) measure \cite{gu2014generating}. 
In contrast, non-pairwise diversity directly measures a set of classifiers using the variance, entropy, or the proportion of individual classifiers that fail on randomly chosen instances~\cite{gu2015multi}.
It includes interrater agreement~\cite{fleiss1981statistical}, Kohavi-Wolpert variance~\cite{kohavi1996bias}, the entropy of the votes~\cite{cunningham2000diversity,shipp2002relationships}, the difficulty index~\cite{hansen1990neural,kuncheva2003diversity}, the generalized diversity~\cite{partridge1997software}, and the coincident failure diversity~\cite{partridge1997software}.
Apart from those, two other measures exist and do not fall into the categories above. 
One is the correlation penalty function, proposed by Liu and Yao~\cite{liu1999ensemble} in their negative correlation learning (NCL) \citep{chen10}. 
It measures the diversity of each member against the entire ensemble.
The other is ambiguity, which measures the average offset of each member against the entire ensemble output~\cite{zenobi2001using}. 
Due to the examined difference or similarity, 
the ranges of diversity vary between different diversity measures \citep{zhou2012ensemble}. 
Moreover, comparing those existing diversity measures is difficult regarding the appropriateness and superiority across them \citep{gu2015multi}. 
The performance of a diversity measure might depend on the context of data and the use of diversity \citep{Tsymbal2005Diversity}. 

\subsection{The Role that Diversity Plays}

Although the crucial role of diversity has been widely accepted, few researchers could tell how diversity works exactly in ensemble methods. 
In the last decade or so, Brown~\cite{brown2009information} claimed that from an information-theoretic perspective, diversity within an ensemble did exist on numerous levels of interaction between the classifiers. 
This work inspired Zhou and Li~\cite{zhou2010multi} to propose that the mutual information should be maximized to minimize the prediction error of an ensemble from the view of multi-information. 
Subsequently, Yu \emph{et al.}~\cite{yu2011diversity} claimed that the diversity among individual learners in a pairwise manner, used in their diversity regularized machine (DRM), could reduce the hypothesis space complexity, which implied that controlling diversity played the role of regularization in ensemble methods. 
Recently, Jiang \emph{et al.}~\cite{jiang2017generalized} extended the classic Ambiguity Decomposition from regression problems to classification problems and proved several fundamental properties of the Ambiguity term. 

\subsection{Ensemble Pruning}

Ensemble pruning, named as ensemble selection or ensemble thinning as well, is proposed to mitigate the shortcoming of ensemble methods that usually require huge memory and processing time regarding the number of individual learners in the ensemble \citep{liu2014ensemble,kokkinos2015confidence,zhang2017ranking,lysiak2014optimal,partalas2008focused,caruana2004ensemble,banfield2005ensemble}. 
It is possible to improve the ensemble generalization performance with a smaller size using ensemble pruning \citep{margineantu1997pruning,zhou2002ensembling}. 
Selecting the sub-ensembles with the best generalization is quite challenging with exponential computational complexity \citep{li2012diversity,martinez2007using}. 
Existing ensemble pruning methods could be categorized basically into three general families: ranking-based, clustering-based, and optimization-based. 
Ranking-based pruning methods sort the learners in the ensemble based on different evaluation function and select the top few of them \citep{tsoumakas2009ensemble,zhang2019two,ahmed2017using}, with a low searching complexity yet theoretical unsoundness \cite{zeng2014constructingBC}; 
Clustering-based pruning methods detect groups of learners making similar predictions and prune each group separately \citep{tsoumakas2004effective,cavalcanti2016combining}; 
Optimization-based pruning methods utilize different objectives to optimize and find the expected subset with satisfactory generalization performance \citep{zhou2003selective,partalas2012study,guo2016novel,qian2015pareto,xia2018maximum,cao2018optimizing}, requiring an optimization searching algorithm to avoid exhaustive search complexity \citep{zeng2014constructingBC}. 
Another partition divides ensemble pruning methods into static and dynamic pruning techniques \citep{06805606}. 
Static pruning techniques focus on identify a small sub-ensemble with good generalization performance, accompanied by expensive computation \cite{zhang2006ensemble,martinez2009analysis,martinez2004aggregation,li2012diversity,partalas2010ensemble,lu2010ensemble};
Dynamic pruning techniques \citep{Hernandez2009Statistical,Martinez2009Statistical, 9060403} determine the number of individual classifiers that need to be queried dynamically for each test instance, reducing the time cost yet not reducing the ensemble's storage requirements \citep{Hernandez2009Statistical,Hernandez2011Inference, fan2002pruning,Wang2003Mining,basilico2011comet}. 
Besides, Jiang \etal{} \citep{07836805} presented a Bayesian framework for ensemble pruning which generated the optimal size of the pruned ensemble first and selected the optimal pruned ensemble accordingly; 
Krawczyk and Wozniak~\cite{07015736} proposed to do ensemble pruning tasks considering the combination rule training simultaneously.

\section{Methodology}
\label{methodology}

In this section, we formally study the measure of diversity using error decomposition of classification ensembles and derive proper theoretical analyses from quantifying and characterizing the relationship between the proposed diversity and generalization performance of the entire ensemble. 
Based on the results of the analyses, two ensemble pruning methods are proposed to construct a pruned sub-ensemble effectively. 

We first introduce the necessary notations used in this paper. 
Let vectors be denoted by bold lowercase letters (e.g., $\mathbf{x}=(x_1,...,x_k)$) as an instance with $k$-tuples of real numbers describing features, and scalars denoted by italic lowercase letters (e.g., $y$) as labels. 
Random variables, (vector) spaces, and subsets of these vector spaces are denoted by sans serif uppercase letters (e.g., $\mathsf{X}$), calligraphic uppercase letters (e.g., $\mathcal{X}$), and roman uppercase letters (e.g., $W$), respectively. 
Sets of vectors are denoted by roman uppercase letters (e.g., $S$) consisting of some instances with their corresponding labels; 
individual classifiers/learners are denoted by functions (e.g., $f(\cdot)$) as the output results of one classifier. 
The symbols $\mathbf{P},\, \mathbf{E}\,,$\, $\mathbb{I}$\,,\, $\mathbb{R}$\,,\, %
and $\mathbb{Z}^+$ denote the probability measure, the expectation of a random variable, the indicator function, the real space, and the positive integer space, respectively. 
The notation $[n]$ represents $\{1,...,n\}$ for clarity. 
The notation $(\mathbf{x},y)\in S$ is formally defined by $(\mathbf{x},y)\in S \Leftrightarrow \exists\, i\in [|S|]: (\mathbf{x}_i,y_i)=(\mathbf{x},y)$ for clarity.

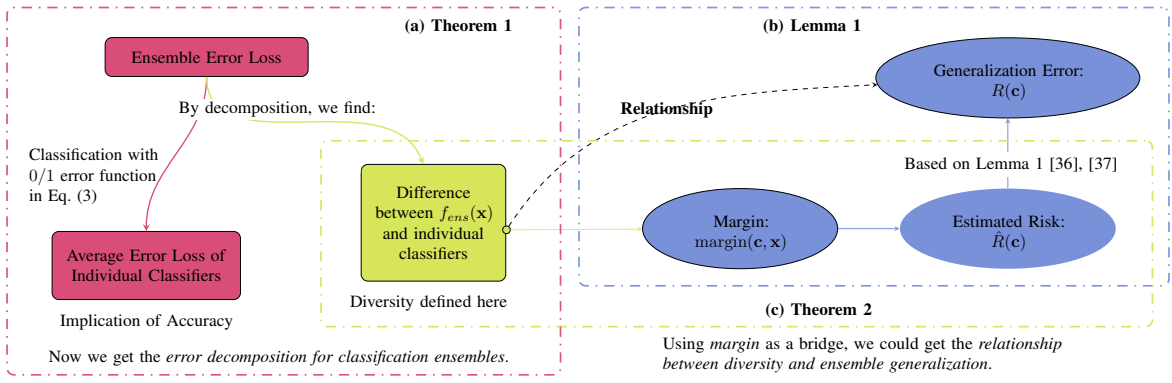
\begin{figure*}[t]
\centering
\scalebox{0.6471}{
\begin{tikzpicture}[%
  block/.style = {draw, fill=blue!30, align=center, anchor=west,
              minimum height=0.65cm, inner sep=0},
  ball/.style = {rectangle, rounded corners, minimum height=0.8cm, draw, align=center, anchor=north, inner sep=1},
  new/.style = {circle, draw, align=center, anchor=north, inner sep=0}]

\node[ball,text width=4.1cm,fill=mypurple] (all) at (3.74,0) {Ensemble Error Loss};
\node[ball,fill=mypurple,minimum height=1.4cm, text width=3.8cm,anchor=base] (pcan) at (2.5,-4.6) {Average Error Loss of Individual Classifiers};
\node[ball,fill=mygreen,minimum height=2.4cm, text width=2.9cm,anchor=base] (pncan) at (8.4,-3.3)
   {Difference between $f_{ens}(\mathbf{x})$ and individual classifiers};

\node[circle,fill=white,text width=0.1cm,anchor=base] (origin) at (0.25,0) {};
\node[new,fill=mygreen, text width=1ex,anchor=base] (diver) at (9.9,-3.96) {};
\node[ellipse,draw,fill=myblue,text width=2.6cm,anchor=base,text centered] (margin) at (14.7, -3.47) {~\\Margin: \\$\margin(\mathbf{c},\mathbf{x})$ \\};
\node[ellipse,fill=myblue,text width=2.9cm,anchor=base,text centered] (estimatedrisk) at (20.2,-3.47) {~\\Estimated Risk: \\$\hat{R}(\mathbf{c})$\\};
\node[ellipse,draw,fill=myblue,text width=3.6cm,anchor=base,text centered] (generalization) at (20.2,-0.37) {~\\Generalization Error: \\$R(\mathbf{c})$\\};

\draw[-stealth,thick,draw=mypurple] (all.south) to [out=270,in=90] (pcan.north);
\draw[-stealth,thick,draw=mygreen] (all.south) to [out=270,in=110] (pncan.100);
\node[rounded corners,minimum height=0.7cm,anchor=north,text width=4.4cm,fill=white,text centered] (tray1) at (2.5,-5.5) {Implication of Accuracy};
\node[rounded corners,minimum height=0.7cm,anchor=north,text width=4.4cm,fill=white,text centered] (tray2) at (8.3,-5.1) {Diversity defined here};

\node[anchor=north,text width=10cm,inner sep=.05cm,align=center,fill=white]
  (why1) at (5.17,-1.27) {By decomposition, we find:\hspace{2em}};
\node[inner sep=0,anchor=east,text width=3.3cm] (note1) at (3.4,-2.87) {
   Classification with \\$0/1$ error function \\in Eq.~\eqref{eq:3}};

\draw[-stealth,text width=3.5cm,color=myblue] (margin) -- (estimatedrisk);
\draw[-stealth,text width=3.5cm,color=myblue] (estimatedrisk) -- (generalization);
\node[inner sep=0.1cm,anchor=west,text width=4.4cm,align=center,fill=white] (based) at (17.96,-2.61) {Based on Lemma~\ref{th:herbrich} \citep{herbrich2001pac,herbrich2002pac}};

\draw[-stealth,draw=black,color=mygreen!80] (diver) -- (margin);
\draw[-stealth,dashed,draw=black,color=black] (diver) to[out=57,in=183] node{\textbf{Relationship}} (generalization);

\node[anchor=north,text width=10cm] (solution) at (5.5,-6.304) {
  Now we get the \emph{error decomposition for classification ensembles}.};
\node[anchor=north,text width=8.7cm] (explain) at (17.46,-6.04) {
  Using \emph{margin} as a bridge, we could get the \emph{relationship between diversity and ensemble generalization}. };

\node[circle,fill=white,text width=0.1cm,anchor=base] (thm1) at (10.4,-6.33) {};
\node[circle,fill=white,text width=0.1cm,anchor=base] (lemma1) at (12,-4.5) {};
\node[circle,fill=white,text width=0.1cm,anchor=base] (lemma2) at (22.9,0) {};
\node[circle,fill=white,text width=0.1cm,anchor=base] (relat1) at (6.7,-2.74) {};
\node[circle,fill=white,text width=0.1cm,anchor=base] (relat2) at (22.56,-5.34) {};

\tikzset{green dotted/.style={draw=mygreen!95!white, line width=1pt,
                             dash pattern=on 1pt off 4pt on 6pt off 4pt,
                             inner sep=4mm, rectangle, rounded corners}};
\tikzset{blue dotted/.style={draw=myblue!95!white, line width=1pt,
                             dash pattern=on 1pt off 4pt on 6pt off 4pt,
                             inner sep=4mm, rectangle, rounded corners}};
\tikzset{violet dotted/.style={draw=mypurple!95!white, line width=1pt,
                               dash pattern=on 1pt off 4pt on 6pt off 4pt,
                               inner sep=4mm, rectangle, rounded corners}};
\node (first dotted box) [blue dotted, fit = (lemma1) (lemma2)] {};
\node (three dotted box) [green dotted, fit = (relat1) (relat2)] {};
\node (second dotted box) [violet dotted, fit = (origin) (thm1)] {};
\node at (first dotted box.north) [below left, inner sep=2.5mm] {\textbf{(b) Lemma~\ref{th:herbrich} }};
\node at (three dotted box.south) [above right, inner sep=2mm] {\hspace{.74em} \textbf{(c) Theorem~\ref{th2} }};
\node at (second dotted box.north) [below right, inner sep=2.5mm] {\hspace{6em} \textbf{(a) Theorem~\ref{th1} }};

\end{tikzpicture}
}
\caption{%
Illustration for the proposed methodology. 
(a) Illustration for the error decomposition for classification ensembles (in Theorem~\ref{th1}). 
(b) Illustration of Lemma~\ref{th:herbrich} \citep{herbrich2001pac,herbrich2002pac}. 
(c) Illustration for the relationship between the proposed diversity and ensemble performance motivated by \citep{herbrich2001pac,herbrich2002pac}. 
Herbrich and Graepel \citep{herbrich2001pac,herbrich2002pac} proposed the relationship between margin and generalization (in Lemma~\ref{th:herbrich}). 
Besides, the proposed diversity in Eq.~\eqref{eq:11} is related to the margin. In this case, we could propose the relationship between the proposed diversity and ensemble performance (in Theorem~\ref{th2}). 
Note that the arrow from A to B means that B is related to A, where arrows in solid lines represent direct relations and that in the dotted line represents the indirect relation. 
}\label{fig:structure}
\end{figure*}

\subsection{Measuring Diversity in Classification Ensembles using Error Decomposition}
\label{propose:measure}

The analysis is built in the PAC (probably approximately correct) framework~\cite{valiant1984theory} that one learning task $\mathcal{D}$ corresponds to a probability distribution over the input-output space $\mathcal{X}\times\mathcal{Y}$. 
An instance from $\mathcal{D}$ is represented as $(\mathbf{x},y)\,,$ where $\mathbf{x} \in \mathbb{R}^k$ is a vector representing features and $y$ is a scalar as a label. 
Suppose this classification task is to use an ensemble that comprises several individual classifiers to approximate a function $f_{true}: \mathcal{X} \mapsto \mathcal{Y} \,,$ and $F=\{f_1,...,f_{|F|}\}$ denotes the set of these individual classifiers. 
The predictions of the individual classifiers are combined by weighted voting
\footnote{
In the spirit of weighted averaging and plurality voting, 
each individual classifier votes for a class with the weight, and the class label receiving the most number of votes is regarded as the output of the ensemble. If there is a tie, the $\Err$ function, Eq.~\eqref{eq:3} is given zero for binary classification problems.}%
, that is, the output of an ensemble is defined as:
\begin{equation}
    \small
	f_{ens}(\mathbf{x}) = \sign\Bigg( \sum_{i\in[|F|]} w_i\cdot f_i(\mathbf{x}) \Bigg)
	\,, \label{eq:1}
\end{equation}%
where $w_i$ is the weight coefficient corresponding to the individual classifier $f_i$\,, satisfying that $\sum_{i\in[|F|]} w_i=1 ,\, \forall w_i\geqslant 0$. 
And $\sum_{i\in[|F|]} w_if_i(\mathbf{x}) =0$ indicates a tie in the combination. 

In this work, we focus on binary classification problems that mean $\mathcal{Y} =\{-1,+1\}$, but these theoretical results could be extended to multi-classification problems (a short example is given in the appendix and more research is left for our future work). 
Assume that there is a training set $S$ with several instances $(\mathbf{x},y)$. 
For one single arbitrary instance $\mathbf{x} \in S \,,\, y$ denotes the target output of this instance, and $f(\mathbf{x})$ is the actual output of the individual classifier $f\in F\,.$
Notice that $y$ and $f(\mathbf{x})$ satisfy that $y,f(\mathbf{x}) \in\mathcal{Y} \,.$ 
If the actual output of the individual classifier is correct according to the target output, obviously $f(\mathbf{x}) \cdot y=+1\,,$ otherwise $f(\mathbf{x}) \cdot y=-1\,,$ where this term is defined as the \emph{margin} of this classifier on the instance,  
\begin{equation}\small
	\margin(f,\mathbf{x}) = f(\mathbf{x}) \cdot y 
	\,.\label{eq:2}
\end{equation}%

As the original error decomposition in regression~\citep{krogh1995neural, brown2005managing, mendes2012ensemble} uses the square loss as the loss function that is not suitable for classification, we employ the $0/1$ error function to adopt the idea of error decomposition for regression ensembles. 
For classification problems, the error function of a classifier $f$ at one single arbitrary instance $\mathbf{x}$ is defined as 
\begin{equation}
\small
\begin{split}
	\Err(f,\mathbf{x})= 
	\begin{cases}
		1 \,, 	& \text{ if } f(\mathbf{x})y = -1 \,;\\
		0.5 \,, & \text{ if } f(\mathbf{x})y = 0  \,;\\
		0 \,,	& \text{ if } f(\mathbf{x})y = 1  \,,
	\end{cases}
\end{split}
\label{eq:3}%
\end{equation}
which is also the discretization of hinge loss function and holds that $\Err(f,\mathbf{x})= -\frac{1}{2}(f(\mathbf{x})y -1)$. 
Therefore, inspired by the work of \cite{brown2005managing,jiang2017generalized}, 
we present the following error decomposition for classification ensembles. 
\begin{thm}[Error decomposition for classification ensembles]\label{th1}
	Assume that we are dealing with binary classification problems. 
	Individual classifiers in an ensemble $F=\{f_1,...,f_{|F|}\}$ have been trained and are combined by weighted voting $f_{ens}(\mathbf{x})=\sign\big(\sum_{i\in[|F|]} w_i f_i(\mathbf{x})\big)$ with $\sum_{i\in[|F|]} w_i=1 ,\, \forall w_i\geqslant 0$\,. 
	Then for the $0/1$ error function, the loss function of the ensemble can be decomposed into 
	\begin{small}
	\begin{align}
	    \Err(f_{ens},\mathbf{x}) =& \sum_{i\in[|F|]} w_i\cdot \Err(f_i,\mathbf{x}) \nonumber\\
		-& \frac{1}{2}\Bigg( \margin(f_{ens},\mathbf{x}) -\sum_{i\in [|F|]}w_i\cdot\margin(f_i,\mathbf{x}) \Bigg) 
		\,.\label{eq:4,th1}
	\end{align}%
	\end{small}%
	Computing the expectation in the data set $S$ yields decomposition of the generalization error as follows,
	\begin{equation}\small
	    \bar{G} = \bar{A} - \bar{D}
	    \,,\label{eq:5,th1}
	\end{equation}
	where
	\begin{subequations}
	\begin{small}
	\begin{align}
		\bar{G}=& \mathbf{E}_{(\mathbf{x},y) \in S}( \Err(f_{ens},\mathbf{x}) ) \,,\\ 
		\bar{A} =& \sum_{i\in [|F|]} w_i\cdot \mathbf{E}_{(\mathbf{x},y) \in S}( \Err(f_i,\mathbf{x}) )  \,,\\
		\bar{D} =& \frac{1}{2} \mathbf{E}_{(\mathbf{x},y) \in S}(\margin(f_{ens},\mathbf{x})) \nonumber\\
		-& \frac{1}{2} \sum_{i\in [|F|]} w_i\cdot  \mathbf{E}_{(\mathbf{x},y) \in S}(\margin(f_i,\mathbf{x}))
		\,.\label{eq:6c,th1}
	\end{align}%
	\end{small}%
	\label{eq:6,th1}%
	\end{subequations}%
\end{thm}

\begin{proof}
	For one single arbitrary instance $\mathbf{x}$\,, the error of an ensemble is defined as 
	\begin{equation}\small
	    \Err(f_{ens},\mathbf{x}) = -\frac{1}{2}\Bigg( \sign\Bigg( \sum_{i\in[|F|]} w_if_i(\mathbf{x}) \Bigg)y -1\Bigg)
	    \,.\label{eq:7}
	\end{equation}
	Inspired by the error decomposition of regression ensembles, the difference between ensemble error and the average error of individual classifiers is 
	\begin{small}
	\begin{align}
	    & \Err(f_{ens},\mathbf{x}) - \sum_{i\in[|F|]} w_i\cdot \Err(f_i,\mathbf{x}) \nonumber\\
	    =& \sum_{i\in[|F|]} w_i\cdot\Big( \Err(f_{ens},\mathbf{x})-\Err(f_i,\mathbf{x}) \Big) \nonumber\\
	    =& -\frac{y}{2}\sum_{i\in[|F|]} w_i\cdot\big( f_{ens}(\mathbf{x})-f_i(\mathbf{x}) \big) 
	    \,,\label{eq:8}
	\end{align}%
	\end{small}%
	and the error decomposition could be rewritten as 
	\begin{small}
	\begin{align}
		\Err(f_{ens},\mathbf{x}) 
		=& \sum_{i\in [|F|]} w_i\cdot \Err(f_i,\mathbf{x}) \nonumber\\
		-& \frac{1}{2} \sum_{i\in [|F|]} w_i\big( f_{ens}(\mathbf{x}) - f_i(\mathbf{x}) \big) y 
		\,.\label{eq:9}
	\end{align}%
	\end{small}%
	The first term, $\sum_{i\in [|F|]} w_i\cdot \Err(f_i,\mathbf{x})$\,, is the weighted average loss of the individuals. 
	The second term is defined as the \emph{diversity} term which measures the difference between $f_{ens}(\mathbf{x})$ and individual classifiers, and it could be rewritten as 
	\begin{small}
	\begin{align}
		& \frac{1}{2}\sum_{i\in [|F|]}w_i\big( f_{ens}(\mathbf{x})-f_i(\mathbf{x}) \big)y \notag\\
		=& \frac{1}{2}\sum_{i\in [|F|]} w_i\big( \margin(f_{ens},\mathbf{x}) - \margin(f_i,\mathbf{x}) \big) 
		\,.\label{eq:10}
	\end{align}%
	\end{small}%
	Note that this term is corresponding to one specific instance $\mathbf{x}$, defined as  
	\begin{equation}\small
		\diver(f_{ens}, \mathbf{x}) = \frac{1}{2}\margin(f_{ens},\mathbf{x}) - 
		\frac{1}{2}\sum_{i\in [|F|]} w_i\cdot \margin(f_i,\mathbf{x})  
		\,.\label{eq:11}
	\end{equation}%
	And then we obtain the form of the error decomposition for one single instance $\mathbf{x}$ as in Eq.~\eqref{eq:4,th1}. 
	Finally, computing the expectation over the data set $S$ will yield the form of the error decomposition for this overall instance set as in Eqs.~(\ref{eq:5,th1}--\ref{eq:6,th1}). 
\end{proof}

Up to now, the diversity measure in Eq.~\eqref{eq:11} is proposed based on the error decomposition for classification ensembles, which would serve as the independent variable in the relationship between the proposed diversity and ensemble performance in Section~\ref{propose:relation}. 
Before that, the properties of the proposed diversity will be investigated in the next subsection.

\subsection{Properties of $\bar{G},\bar{A},\bar{D}$}
\label{propose:property}

In this section, we analyze $\bar{G},\bar{A},\bar{D}$ in Eq.~\eqref{eq:5,th1} of Theorem~\ref{th1} to further explore the properties of these three terms. 
\begin{corol}[$\bar{G}\,,\bar{A}\,,\bar{D}$ only depend on $\diver(f_{ens},\mathbf{x})$]\label{th3}
	Assume that we are dealing with binary classification problems. 
	Individual classifiers in an ensemble $F=\{f_1,...,f_{|F|}\}$ have been trained and are combined by weighted voting $f_{ens}(\mathbf{x})=\sign(\sum_{i\in [|F|]} w_i\cdot f_i(\mathbf{x}))$ with $\sum_{i\in [|F|]} w_i=1,\, \forall w_i\geqslant 0$\,. 
	Then the generalization error $\bar{G}$, accuracy $\bar{A}$ and diversity $\bar{D}$ of this ensemble are all dependent on $\diver(f_{ens},\mathbf{x})$, that is, 
	\begin{subequations}
	\begin{small}
	\begin{align}
		\bar{G}=& \frac{1}{2}\Big(1-\mathbf{E}_{(\mathbf{x},y)\in S}\big( \sign(\lambda-2\diver(f_{ens},\mathbf{x})) \big)\Big) \label{eq:12a,corol}\,,\\
		\bar{A}=& \frac{1}{2}\Big(1-\mathbf{E}_{(\mathbf{x},y)\in S}\big( \lambda-2\diver(f_{ens},\mathbf{x}) \big)\Big) \label{eq:12b,corol}\,,\\
		\bar{D}=& \mathbf{E}_{(\mathbf{x},y)\in S}\big( \diver(f_{ens},\mathbf{x}) \big)
		\label{eq:12c,corol}\,,
	\end{align}%
	\end{small}%
	\label{eq:12,corol}%
	\end{subequations}%
	where 
	\begin{equation}\small
	\begin{split}
		\lambda = 
		\begin{cases}
			1 \,, & \text{ if } \diver(f_{ens},\mathbf{x}) \in (0,\frac{1}{2}) \,; \\
			0 \,, & \text{ if } \diver(f_{ens},\mathbf{x}) = 0 \,; \\
			-1\,, & \text{ if } \diver(f_{ens},\mathbf{x}) \in (-\frac{1}{2},0) \,.
		\end{cases}%
	\end{split}%
	\label{eq:13,lambda}%
	\end{equation}%
\end{corol}
\begin{proof}
	The error of one individual classifier is 
	\begin{small}
	\begin{align}
	    \Err(f,\mathbf{x}) =& -\frac{1}{2}\big(f(\mathbf{x})y-1\big) \nonumber\\
	    =& -\frac{1}{2}\big(\margin(f,\mathbf{x}) -1\big) \,,
	\end{align}%
	\end{small}%
	as described previously, therefore, according to Theorem~\ref{th1}, we could obtain that 
	\begin{subequations}
	\begin{small}
	\begin{align}
		\bar{G} 
		=& \frac{1}{2}\Big(1- \mathbf{E}_{(\mathbf{x},y)\in S}\big( \margin(f_{ens},\mathbf{x}) \big) \Big) \,,\\
		\bar{A} 
		=& \frac{1}{2}\bigg( 1-\mathbf{E}_{(\mathbf{x},y)\in S} \Big( \sum_{i\in [|F|]}w_i\cdot \margin(f_i,\mathbf{x}) \Big) \bigg) \,,\\
		\bar{D}
		=& \frac{1}{2}\sum_{i\in [|F|]} w_i\mathbf{E}_{(\mathbf{x},y)\in S}\big( \margin(f_{ens},\mathbf{x}) - \margin(f_i,\mathbf{x} ) \big) \,.
	\end{align}%
	\end{small}%
	\label{eq:15}
	\end{subequations}%
	On the other side, according to Eq.~\eqref{eq:11}, we could obtain that 
	\begin{small}
	\begin{align}
		& \diver(f_{ens},\mathbf{x}) \nonumber\\
		=& \frac{1}{2}\sign\Bigg( \sum_{i\in [|F|]}w_if_i(\mathbf{x}) \Bigg)y -\frac{1}{2}\sum_{i\in [|F|]}w_i\margin(f_i,\mathbf{x}) \nonumber\\
		=& \frac{1}{2}  \sign\Bigg( \sum_{i\in [|F|]}w_i \margin(f_i,\mathbf{x}) \Bigg) -\frac{1}{2} \sum_{i\in [|F|]}w_i \margin(f_i,\mathbf{x}) 
		\,,\label{eq:16}
	\end{align}%
	\end{small}%
	then define that
	\begin{equation}\small
    	\overline{\margin}(f_{ens},\mathbf{x}) \defineq 
	    \sum_{i\in [|F|]} w_i\cdot \margin(f_i,\mathbf{x}) 
	    \,,\label{eq:17}
	\end{equation}
	and consequently obtain that for clarity, 
	\begin{equation}\small
		\overline{\margin}(f_{ens},\mathbf{x}) = \lambda -2\diver(f_{ens},\mathbf{x}) 
		\,,\label{eq:18}
	\end{equation}
	where $\lambda$ is defined in Eq.~\eqref{eq:13,lambda}. 
	Finally, we could obtain the Eqs.~\eqref{eq:12a,corol}--\eqref{eq:12c,corol} as described in Corollary~\ref{th3}. 
\end{proof}

\subsection{Relationship Between Diversity and Ensemble Performance}
\label{propose:relation}

In this subsection, we intend to explain whether and how diversity affects the generalization performance of classification ensembles. 
Herbrich and Graepel~\cite{herbrich2001pac,herbrich2002pac} presented that the relationship between generalization and margin did exist. 
Based on their work and the relationship between the proposed diversity and margin, we propose the relationship between the proposed diversity and generalization in the PAC framework, as shown in Figure~\ref{fig:structure}.

Given the input/feature space $\mathcal{X}\,,$ the output/label space $\mathcal{Y}\,,$ and the training set $S$ as described above, 
any instance $(\mathbf{x},y) \sim (\mathsf{X,Y}) \in (\mathcal{X,Y})$ in $S$ is drawn independent and identically distributed (i.i.d.) according to a certain unknown probability measure $\mathbf{P}_{\mathsf{Y|X}} \mathbf{P}_{\mathsf{X}}\,,$ and we only consider linear classifiers as ensembles just like $f_{ens}(\cdot)$ in Eq.~\eqref{eq:1}, \ie{}
\begin{equation}\small
	\mathcal{F} = \{
	\mathbf{x}\mapsto \sign(\langle \mathbf{w},\phi(\mathbf{x}) \rangle) 
	\,|\, \mathbf{w}\in\mathcal{W}
	\}  \,,\label{eq:19}
\end{equation}
where $\phi(\mathbf{x}) = [f_1(\mathbf{x}), ..., f_{|F|}(\mathbf{x})] ^{\mathsf{T}}$ and $\mathbf{w}=[w_1,...,w_{|F|}]^\mathsf{T}$. 
Notice that $\sum_{i\in [|F|]}w_i=1$ leads to a one-to-one correspondence of hypotheses $f_{ens}(\mathbf{x}, \mathbf{w}) \in\mathcal{F}$ to their parameters $\mathbf{w}\in\mathcal{W}$. 
Assume that the existence of a ``true'' hypothesis $\mathbf{w}^* \in\mathcal{W}$ labeling the data, leading to a PAC-likelihood, 
\begin{equation}\small
	\mathbf{P}_{ \mathsf{Y}| \mathsf{X}= \mathbf{x} }(y) = 
	\mathbb{I}\big( y=\sign(\langle \mathbf{w}^*,\phi(\mathbf{x}) \rangle) \big)
	\,.\label{eq:20}
\end{equation}
There exists a version space $V(S) \subseteq \mathcal{W}$ because of the existence of $\mathbf{w}^*$, 
\begin{equation}\small 
	V(S) = \{ \mathbf{w}\in\mathcal{W} \,|\, \forall (\mathbf{x},y) \in S\,,\; \sign(\langle \mathbf{w},\phi(\mathbf{x}) \rangle) =y \} 
	\,.\label{eq:21}
\end{equation}
The true risk $R(\mathbf{w})$ of consistent hypotheses $\mathbf{w}\in V(S)$ is 
\begin{equation}\small
	R(\mathbf{w}) = \mathbf{E}_{\mathsf{XY}}\big[ \mathbb{I}\big( \sign(\langle \mathbf{w},\phi(\mathsf{X}) \rangle) \neq \mathsf{Y} \big)\big] 
	\,,\label{eq:22}
\end{equation}
and a concept of the \emph{margin} $\gamma(\mathbf{w})$ of an ensemble $\mathbf{w}$ is introduced~\cite{herbrich2001pac, herbrich2002pac} as 
\begin{equation}\small
    \gamma(\mathbf{w}) = \min_{(\mathbf{x},y) \in S} y\langle \mathbf{w},\phi(\mathbf{x}) \rangle
    \,,\label{eq:23}
\end{equation}
since all ensemble classifiers, $\mathbf{w}\in V(S)$ are indistinguishable in terms of error rate on the given training set $S\,.$
Herbrich and Graepel~\cite{herbrich2002pac} presented that the generalization error $R(\mathbf{w})$ of an ensemble was bounded from above in terms of the margin $\gamma(\mathbf{w})$, as shown in Lemma~\ref{th:herbrich}. 
\begin{lemma}[Relationship between margin and generalization~\cite{herbrich2002pac}]\label{th:herbrich}
	For any probability measures $\mathbf{P}_{\mathsf{X}}$ such as $\mathbf{P}_{\mathsf{X}} (\Vert \phi(\mathbf{x}) \Vert \leqslant \delta) =1$ and for any $\xi \in (0,1] \,,$ 
	with probability at least $(1-\xi)$ over the random draw of the training instances $S\in (\mathcal{X},\mathcal{Y})^{|S|} \,,$ 
	for any arbitrary consistent ensemble classifier $\mathbf{w}\in V(S)$ with a positive margin $\gamma(\mathbf{w}) > \sqrt{32\delta^2/|S|} \,,$ 
	the generalization error $R(\mathbf{w})$ could be bounded from above by 
	\begin{equation}\small 
    	R(\mathbf{w}) \leqslant 
	    \frac{2}{|S|} \bigg( \kappa(\mathbf{w})\log_2\Big( \frac{8e|S|}{\kappa(\mathbf{w})} \Big) \log_2(32|S|) + \log_2\Big( \frac{2|S|}{\xi} \Big) \bigg)
    	\,,\label{eq:24,herbrich}
	\end{equation}
	where $\kappa(\mathbf{w}) = \Big\lceil\big( \frac{8\delta}{\gamma(\mathbf{w})} \big)^2\Big\rceil \,.$
\end{lemma}
Subsequently, inspired by the relationship between margin $\gamma(\mathbf{w})$ and generalization error $R(\mathbf{w})$ in Lemma~\ref{th:herbrich}~\citep{herbrich2002pac}, we present the following relationship between the proposed diversity and generalization. 
It is worth mentioning that the margin $\gamma(\mathbf{w})$ could be linked with the proposed diversity $\diver(f_{ens}, \mathbf{x})$ in Eq.~\eqref{eq:11}, which exactly inspired us to pursue the relationship between the proposed diversity and the generalization risk. 
To simplify the analysis, we only consider the most important item in Eq.~\eqref{eq:24,herbrich}, \ie{} $\kappa(\mathbf{w})\log_2\big( \frac{8e|S|}{\kappa(\mathbf{w})} \big) \,,$ described as $\hat{R}(\mathbf{w}) \,.$

\begin{figure}[tb]
\centering 
\subfloat[]{\label{prop:a}\centering
	\includegraphics[scale=.291]{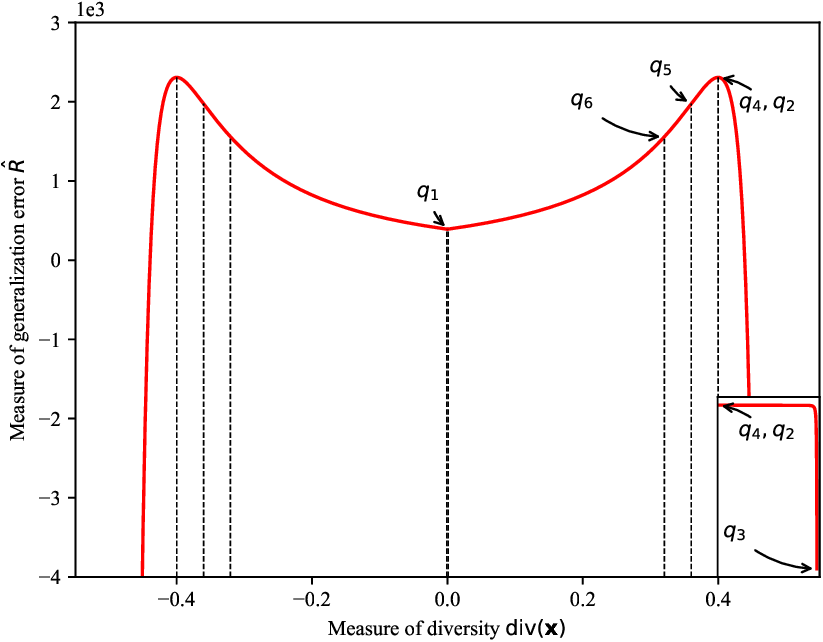}
}%
\hspace{0.8em}
\subfloat[]{\label{prop:b}\centering
	\includegraphics[scale=.291]{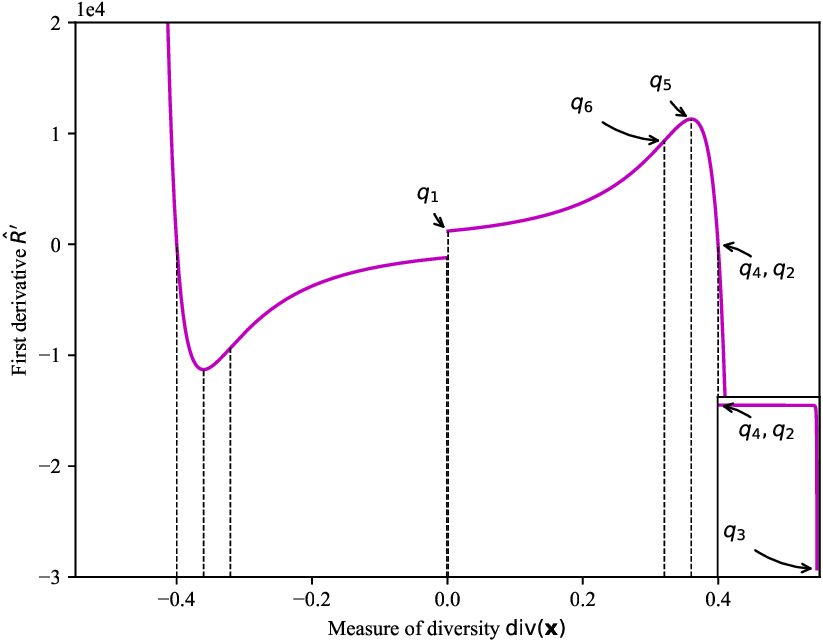}
}%
\caption{
Illustration of the estimator of generalization error and its first derivative, impacted by our proposed measure of diversity. Notice that these endpoints ($q_4,q_5$, and $q_6$) in this figure refer to those points on the horizontal axis with the same abscissas correspondingly in fact. In this graph, parameters' values are specialized as follows: $\delta =1$, $\varepsilon =0.01$, $|S|=200$. 
(a) Relationship between diversity and the ensemble generalization. 
(b) Relationship between diversity and the first derivative of generalization. 
}\label{proposition}
\end{figure}

\begin{thm}[Relationship between the proposed diversity and generalization]\label{th2}
	For any probability measures $\mathbf{P}_{\mathsf{X}}$ such as $\mathbf{P}_{\mathsf{X}} (\Vert \phi(\mathbf{x}) \Vert \leqslant \delta) =1$ and for any $\xi \in (0,1]$\,, 
	the random draw of the training instances $S\in (\mathcal{X,Y})^{|S|}$ is given with probability $\varepsilon$ by noise disturbance. 
	Then for any arbitrary consistent ensemble classifier $\mathbf{w}\in V(S)$\,, with probability at least $(1-\xi)$\,, 
	the variation tendency of the upper bound of generalization error $R(\mathbf{w})$ is the same as that of $\hat{R}(\mathbf{w})$\,, \ie{}%
	\begin{equation}\small
		\hat{R}(\mathbf{w}) = \Big( \frac{8\delta}{\gamma(\mathbf{w})} \Big)^2 \log_2 \bigg( 8e|S|\Big( \frac{\gamma(\mathbf{w})}{8\delta} \Big)^2 \bigg) 
		\,,\label{eq:25}
	\end{equation}
	\begin{equation}\small
		\gamma(\mathbf{w}) = \min_{(\mathbf{x},y) \in S} \big(1-2\varepsilon\big)\big(\lambda-2\diver(f_{ens},\mathbf{x})\big) 
		\,,\label{eq:26}
	\end{equation}
	where $\lambda$ is defined as Eq.~\eqref{eq:13,lambda}.
\end{thm}

\begin{proof} 
	Due to the existence of noise in real data, for any arbitrary instance $\mathbf{x}$ in $S$, 
	let $z$ be the observed label of this instance, and let $y$ be the true unknown label correspondingly. 
	Thus, the error rate of the training set $S$ is recorded as $\mathbf{P}(z\neq y\,|\, \mathbf{x}\in S) \defineq\varepsilon$, 
	and the error of a classifier $f$ is recorded as $\mathbf{P}(f(\mathbf{x}) \neq z\,|\, \mathbf{x}\in S) \defineq\theta$. 
	Since the training set is affected by noise and other factors, we believe that $\theta \geqslant \varepsilon$. 
	Then we use the observable $z$ to estimate the value of $y$, \ie{} $\hat{y}= (1-\varepsilon)z+\varepsilon(-z) =(1-2\varepsilon)z$. 
	Thus, for a given training set $S=\{(\mathbf{x}_i,z_i) \,|\, i\in[|S|]\}$, individual classifiers are trained to constitute an ensemble, \ie{} the set $F=\{f_1,...,f_{|F|}\}$. 
	The margin of one individual classifier $f_j$ on the instance $\mathbf{x}_i$ is 
	\begin{equation}\small
    	\margin(f_j, \mathbf{x}_i) = f_j(\mathbf{x}_i) z_i 
    	\,.\label{eq:27}
	\end{equation}
	Describe $\overline{\margin}(f_{ens},\mathbf{x}_i)$ as in Eq.~\eqref{eq:17}, and then according to Eq.~\eqref{eq:18}, the margin of the ensemble is obtained 
	\begin{equation}\small
	    \gamma(\mathbf{w}) = \min_{i\in [|S|]} \big( 1-2\varepsilon \big)\big( \lambda-2\diver(f_{ens},\mathbf{x}_i) \big) 
	    \,,\label{eq:28}
	\end{equation}
	based on Eq.~\eqref{eq:23},
	which is mainly reflecting the difference among different members of the ensemble, \ie{} diversity.
	Let $\mathbf{x}^*$ be the very instance where the ensemble reaches the minimum of the margin $\gamma(\mathbf{w}) \,,$ that is, 
	\begin{equation}\small
	    \gamma(\mathbf{w}) = \big( 1-2\varepsilon \big)\big( \lambda-2\diver(f_{ens},\mathbf{x}^*) \big) 
	    \,,\label{eq:29}
	\end{equation}
	then $\kappa(\mathbf{w})$ is obtained based on Eq.~\eqref{eq:24,herbrich}. 
	To simplify the analysis, we only consider the most important item in Eq.~\eqref{eq:24,herbrich}, 
	\ie{} $\kappa(\mathbf{w})\log_2\big( \frac{8e|S|}{\kappa(\mathbf{w})} \big) \,,$ described as $\hat{R}(\mathbf{w}) \,,$ that is,
	\begin{footnotesize}
	\begin{align}
        \hat{R}(\mathbf{w}) \defineq& \kappa(\mathbf{w})\log_2\bigg( \frac{8e|S|}{\kappa(\mathbf{w})} \bigg) \nonumber\\
	    \approx& 
	    \bigg( \frac{8\delta}{(1-2\varepsilon)(\lambda-2\diver(f_{ens},\mathbf{x}^*))} \bigg)^2  \nonumber\\
	    &\cdot \log_2\Bigg( 8e|S|\bigg( \frac{(1-2\varepsilon)(\lambda-2\diver(f_{ens},\mathbf{x}^*))}{8\delta} \bigg)^2 \Bigg)
	    \,,\label{eq:30,derivative0}
	\end{align}%
	\end{footnotesize}%
	with the same variation tendency as $R(\mathbf{w})$\,.
\end{proof}

According to Theorem~\ref{th2}, the generalization error of the ensemble could be quantified by the diversity. 
Eq.~\eqref{eq:30,derivative0} reflects the relationship of individual classifiers $\diver(f_{ens}, \mathbf{x}^*) \in(-\frac{1}{2},\frac{1}{2})$ and the performance of generalization error of the ensemble $R(\mathbf{w})$. 
Therefore, the first and the second derivative\footnote{%
The first derivative of a function $f(x)$, written as $f'(x)$, is the slope of the tangent line to the function at the point $x$. To put this in non-graphical terms, the first derivative tells us whether a function is increasing or decreasing, and by how much it is increasing or decreasing. 
The second derivative of a function is the derivative of that function's derivative, written as $f''(x)$. 
While the first derivative can tell us if the function is increasing or decreasing, the second derivative tells us if the first derivative is increasing or decreasing~\cite{derivative_lecture}.
} of the estimated risk $\hat{R}(\mathbf{w})$ could be obtained by taking the derivative of $\hat{R}(\mathbf{w})$ on $\diver(f_{ens},\mathbf{x}^*)$, \ie{}
\begin{footnotesize}
\begin{align}
	\hat{R}^{\prime} =&	 4\left(\frac{8\delta}{1-2\varepsilon}\right)^2 \frac{ 1 }{(\lambda-2\diver(f_{ens},\mathbf{x}^*))^3}  \notag\\
	&  \cdot \log_{2}\left(8|S|\left(\frac{(1-2\varepsilon)(\lambda-2\diver(f_{ens},\mathbf{x}^*))}{8\delta}\right)^2\right)  
	\,,\label{eq:31,derivative1}\\
	\hat{R}^{\prime\prime} =&	\frac{8}{\ln 2} \left(\frac{8\delta}{1-2\varepsilon}\right)^2 \frac{ 1 }{(\lambda-2\diver(f_{ens},\mathbf{x}^*))^4}  \notag\\
	\cdot & \left( 3\ln\left(8|S| \left(\frac{(1-2\varepsilon)(\lambda-2\diver(f_{ens},\mathbf{x}^*))}{8\delta}\right)^2 \right) -2\right)  
	\,.\label{eq:32,derivative2}
\end{align}%
\end{footnotesize}%
We are already aware that $\diver(f_{ens},\mathbf{x}^*) \in(-\frac{1}{2},\frac{1}{2})$. 
Thus, the graph of the upper bound of the generalization error $\hat{R}(\mathbf{w})$ is symmetrical about the vertical axis, as shown in Figure~\ref{proposition}. 
And then we take $\diver(f_{ens},\mathbf{x}^*) \in(0,\frac{1}{2})$ for example to analyze the effect of diversity on the generalization error in the ensemble.

\begin{table}[tbhp]
\centering
\caption{
Monotone intervals. 
The first column is diversity $\diver(f_{ens},\mathbf{x}^*)$. 
The second and the third columns are the estimated risk $\hat{R}(\mathbf{w})$ and its first derivative, respectively. 
Note that the estimated risk reflects the generalization error $R(\mathbf{w})$ of the ensemble $f_{ens}$. 
The fourth and the fifth columns are the variation of the estimated risk and that of its first derivative, respectively.
}\label{monotone_intervals}
\scalebox{1}{
	\begin{tabular}{cllllll}
		\toprule
		$\diver(f_{ens},\mathbf{x}^*)$ & $\hat{R}(\mathbf{w})$ & $\hat{R}'(\mathbf{w})$ & $\Delta\hat{R}$ & $\Delta\hat{R}'$ \\
		\midrule
		$(-q_3,-q_2)$&$\nearrow$ convex &$\searrow$ concave&smaller&larger \\
		$(-q_2,-q_5)$&$\searrow$ convex &$\searrow$ concave&smaller&larger \\
		$(-q_5,-q_6)$&$\searrow$ concave&$\nearrow$ concave&larger &larger \\
		$(-q_6,-q_1)$&$\searrow$ concave&$\nearrow$ convex &larger &smaller\\
		$(q_1,q_6)$  &$\nearrow$ concave&$\nearrow$ concave&larger &larger \\
		$(q_6,q_5)$  &$\nearrow$ concave&$\nearrow$ convex &larger &smaller\\
		$(q_5,q_2)$  &$\nearrow$ convex &$\searrow$ convex &smaller&smaller\\
		$(q_2,q_3)$  &$\searrow$ convex &$\searrow$ convex &smaller&smaller\\
		\bottomrule
	\end{tabular}
}
\end{table}

As the relationship between the proposed diversity $\diver(f_{ens},\mathbf{x}^*)$ and the generalization error $\hat{R}(\mathbf{w})$ in the ensemble is Eq.~\eqref{eq:30,derivative0}, the monotonicity and concavity of function could be analyzed based on its first derivative and second derivative. 
On the positive abscissa axis, let $q_1$ and $q_3$ be the left endpoint and the right endpoint of the range, respectively. 
The function's inflecion point $q_2$ is found by making $\hat{R}^{\prime}(\mathbf{w}) =0$\,, also remarked as $q_4$\,. 
Thus the monotone increasing interval is $(q_1,q_2)$\,, and the monotone diminishing interval is $(q_2,q_3)$\,. 
Then the stagnation point of the first derivative $q_5$ is found by making $\hat{R}'^{\prime}(\mathbf{w}) =0$\,, to analyze the concavity of the function $\hat{R}(\mathbf{w})$\,; 
the stagnation point of the second derivative $q_6$ is found by making $\hat{R}'''(\mathbf{w}) =0$\,, to analyze the concavity of the function $\hat{R}'(\mathbf{w})$\,. 
Their values are 
\begin{subequations}
\begin{footnotesize}
\begin{align}
    q_1= \varepsilon\,,\;\; q_3=& \frac{1}{2}\bigg( 1-\frac{\varepsilon}{1-2\varepsilon} \bigg) \,,
	\quad\quad\quad\quad\,\,\,\\
	q_4=q_2=& \frac{1}{2}\Bigg( 1-\frac{\delta}{1-2\varepsilon} \sqrt{\frac{8}{|S|}} \,\Bigg) \,,
	\quad\quad\quad\,\,\\
	q_5=& \frac{1}{2}\Bigg( 1-\frac{\delta}{1-2\varepsilon} \sqrt{\frac{8}{|S|} e^{\frac{2}{3}}} \,\Bigg) \,,\\
	q_6=& \frac{1}{2}\Bigg( 1-\frac{\delta}{1-2\varepsilon} \sqrt{\frac{8}{|S|} e^{\frac{7}{6}}} \,\Bigg) \,,
\end{align}%
\end{footnotesize}%
\label{eq:33,endpoint}%
\end{subequations}%
respectively, where an implied condition exists, \ie{}
\begin{equation}
\footnotesize
	\varepsilon\leqslant  
	\frac{\delta}{1-2\varepsilon} \sqrt{\frac{8}{|S|}}  
	\leqslant 1-2\varepsilon  
	\,.\label{eq:34,imply}
\end{equation}
Note that $q_1<q_2<q_3$ and $q_6<q_5<q_4=q_2\,.$ 
Then the domain of diversity $\diver(f_{ens},\mathbf{x}^*)$ could be divided into eight intervals, as shown in Table~\ref{monotone_intervals}. 
Up to now, we could analyze in detail the effect of diversity on generalization error within different intervals. 
\begin{enumerate}[(1)]%
	\item  When $\diver(f_{ens},\mathbf{x}^*)=0$ in this paper, we only consider two cases: ``all individual classifiers give the right result of the instance $\mathbf{x}^*$'s label;'' ``all individual classifiers give the wrong result of the instance $\mathbf{x}^*$'s label.'' We do not consider the situation of a tie. 
	\item In real situations, $\diver(f_{ens},\mathbf{x}^*)$ is negative in most cases which means that ``the ensemble classifier has misclassified the instance that corresponds to the selected $\diver(f_{ens},\mathbf{x}^*)$.'' Notice that in the finite experiments, when the value of $|S|$ satisfies Eq.~\eqref{eq:34,imply}, %
	it is rare that $\diver(f_{ens},\mathbf{x}^*)$ is larger than zero. 
	If $\diver(f_{ens},\mathbf{x}^*)<0\,,$ increasing diversity will improve the ensemble's performance, which is the reason why we should increase diversity in that case. 
	\item When $\diver(f_{ens},\mathbf{x}^*)$ is located in the positive abscissa axis, the ensemble classifies the corresponding instance correctly. 
	In the range near zero, individual classifiers that classify the instance correctly take great advantage, and the closer $\diver(f_{ens},\mathbf{x}^*)$ is to zero, the greater their advantage is. 
	In the range near $1/2\,,$ individual classifiers that classify the instance correctly take little advantage, and the closer $\diver(f_{ens},\mathbf{x}^*)$ is to $1/2\,,$ the less their advantage is. 
	In this case, diversity needs to be decreased to increase the generalization performance of the ensemble, with the basic idea of keeping the ensemble classifying the corresponding instance correctly. 
	In summary, increasing diversity will reduce the advantage of the individuals that classify the instance correctly, which should be avoided. 
	\item When $\diver(f_{ens},\mathbf{x}^*)$ is located in the negative axis, the ensemble classifies the instance incorrectly. Similar results could be analogized. 
\end{enumerate}
Based on these analyses, the relationship between the proposed diversity and generalization error varies when diversity falls in different ranges. 
Generally speaking, there are two specific ranges that need to be paid attention to particularly, i.e., $(-q_5,-q_6)$ and $(-q_6,-q_1)\,,$ according to Table~\ref{monotone_intervals}. 
In these two ranges, more diversity would lead to better generalization, which is why it is necessary to increase $\diver(f_{ens},\mathbf{x}^*)$ to get a more effective ensemble classifier. 
However, in other ranges, it is not worth increasing diversity, and the ensemble could remain the same to continue the learning process.

\subsection{\pdfull{} and One Ensemble Pruning Framework}
\label{propose:pruning}

Inspired by the relationship between the proposed diversity and generalization of classification ensembles, we propose an ensemble pruning method, named as ``\emph{\pdfull{} (}\pdabbr{}\emph{)},'' presented in Algorithm~\ref{algorithm}. 
It aims to prune an ensemble with rare performance degradation, with the basic idea of utilizing diversity and accuracy simultaneously.

\begin{algorithm}[tbhp]
\centering\small%
\caption{\small \pdfull{} (\pdabbr{})}%
\label{algorithm}
\begin{algorithmic}[1]
	\REQUIRE %
	A training set $S= \{(\mathbf{x}_i,y_i) |\, i\in[|S|]\}$, 
	an original ensemble $F= \{f_j |\, j\in[|F|]\}$, 
	and the threshold $\alpha$. 
	\ENSURE The pruned sub-ensemble $H \,(H \subset F)$. 
	\STATE $H = \emptyset$. 
	\REPEAT
	    \STATE Search for the specific data instance $(\mathbf{x},y)$ which satisfies the search criterion Eq.~\eqref{eq:13,lambda}. \label{alg:3} 
		\STATE Sort classifiers in $F$ that classify this instance correctly in ascending order according to the accuracy performance. \label{alg:4} 
		\STATE Move the top one $f(\cdot)$ in the previous step from $F$ to $H$\,. \label{alg:5}
	\UNTIL{The termination condition is satisfied.} 
\end{algorithmic}
\end{algorithm}

\begin{figure*}[htbp]
\centering
\begin{minipage}{\linewidth}
\centering%
\subfloat[]{\centering
    \includegraphics[scale=0.4124]{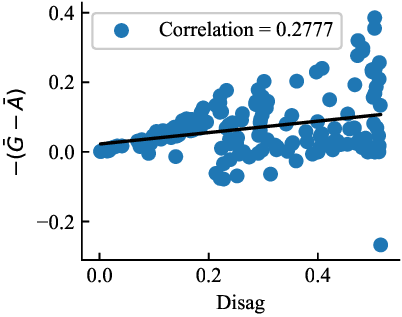}}
\subfloat[]{\centering
    \includegraphics[scale=0.4124]{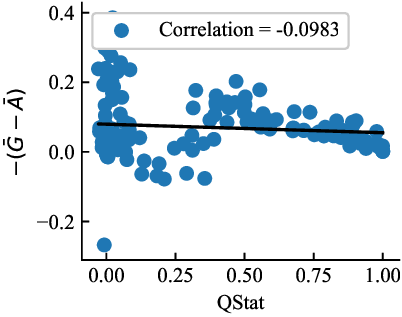}}
\subfloat[]{\centering
    \includegraphics[scale=0.4124]{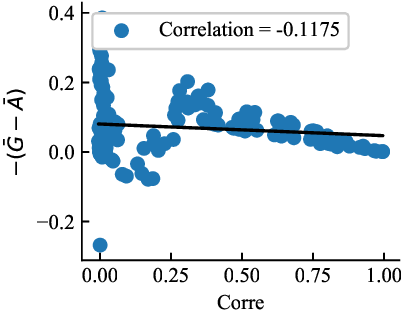}}
\subfloat[]{\centering
    \includegraphics[scale=0.4124]{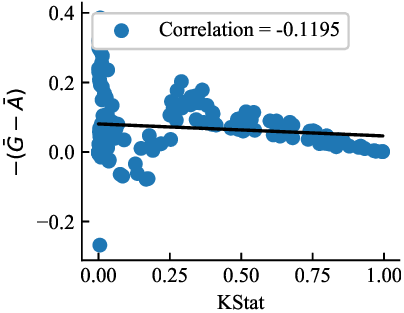}}
\subfloat[]{\centering
    \includegraphics[scale=0.4124]{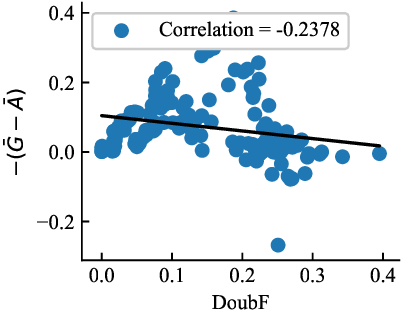}}
\subfloat[]{\centering
    \includegraphics[scale=0.4124]{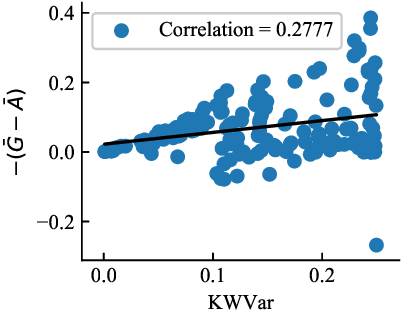}}
\\
\subfloat[]{\centering
    \includegraphics[scale=0.4124]{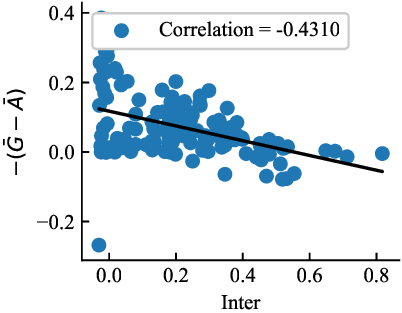}}
\subfloat[]{\centering
    \includegraphics[scale=0.4124]{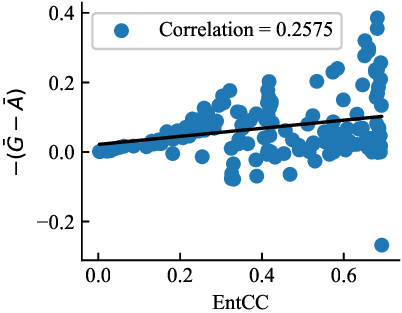}}
\subfloat[]{\centering
    \includegraphics[scale=0.4124]{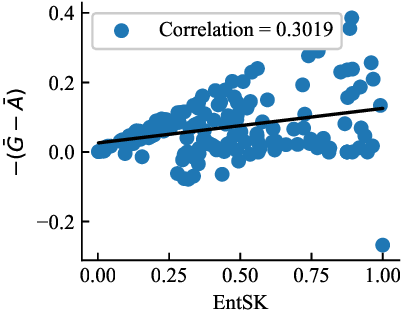}}
\subfloat[]{\centering
    \includegraphics[scale=0.4124]{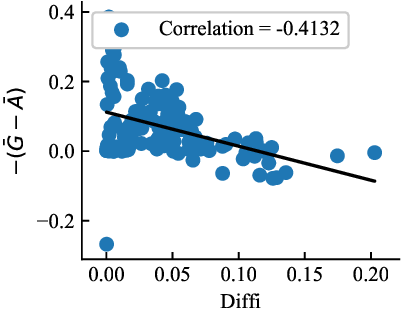}}
\subfloat[]{\centering
    \includegraphics[scale=0.4124]{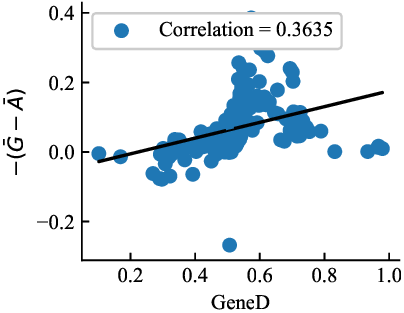}}
\subfloat[]{\centering
    \includegraphics[scale=0.4124]{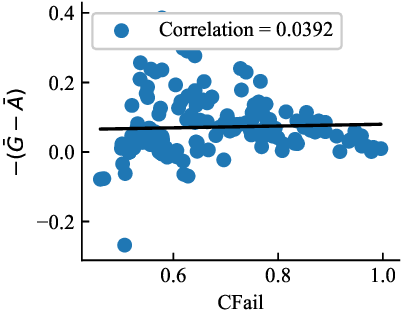}}
\caption{%
Relationship of the loss difference $-(\bar{G}-\bar{A})$ in Eq.~\eqref{eq:5,th1} and different diversity measures, using Bagging to conduct ensemble classifiers and MLPs as individual classifiers. 
(a) Disagreement measure \citep{skalak1996sources,ho1998random}; 
(b) $Q$-statistic~\citep{yule1900vii}; 
(c) Correlation coefficient~\citep{sneath1973numerical}; 
(d) $\kappa$-statistic~\citep{cohen1960coefficient}; 
(e) Double-fault~\citep{giacinto2001design}; 
(f) Kohavi-Wolpert variance~\citep{kohavi1996bias}; 
(g) Interrater agreement~\citep{fleiss1981statistical}; 
(h--i) The entropy of the votes~\citep{cunningham2000diversity,shipp2002relationships}; 
(j) The difficulty index~\citep{hansen1990neural,kuncheva2003diversity};
(k) The generalized diversity~\citep{partridge1997software};
(l) The coincident failure diversity~\citep{partridge1997software}. 
}\label{fig:decom:Bag,NN}
\end{minipage}
\begin{minipage}{\linewidth}
\centering %
\subfloat[]{\centering
    \includegraphics[scale=0.4124]{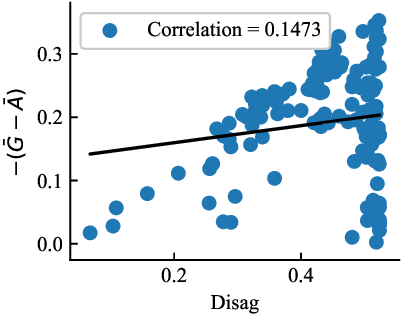}}
\subfloat[]{\centering
    \includegraphics[scale=0.4124]{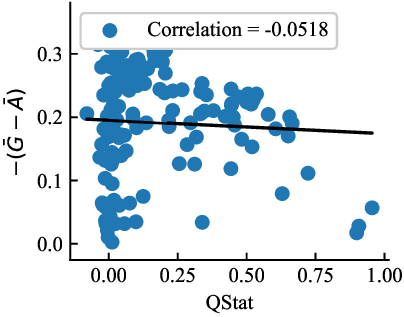}}
\subfloat[]{\centering
    \includegraphics[scale=0.4124]{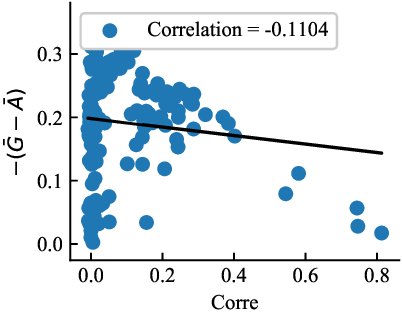}}
\subfloat[]{\centering
    \includegraphics[scale=0.4124]{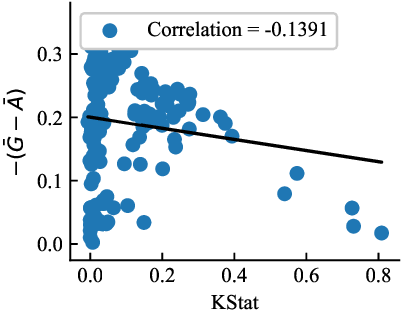}}
\subfloat[]{\centering
    \includegraphics[scale=0.4124]{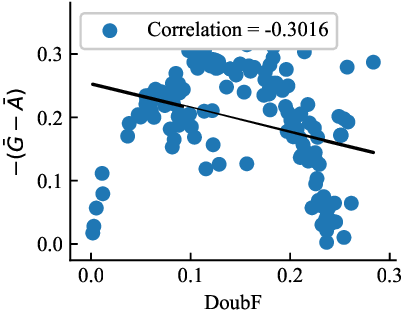}}
\subfloat[]{\centering
    \includegraphics[scale=0.4124]{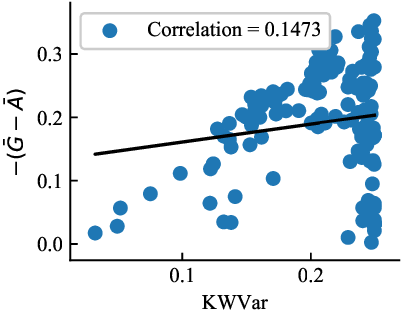}}
\\
\subfloat[]{\centering
    \includegraphics[scale=0.4124]{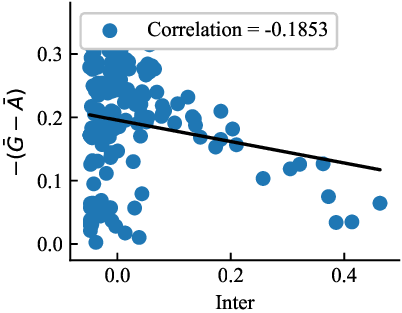}}
\subfloat[]{\centering
    \includegraphics[scale=0.4124]{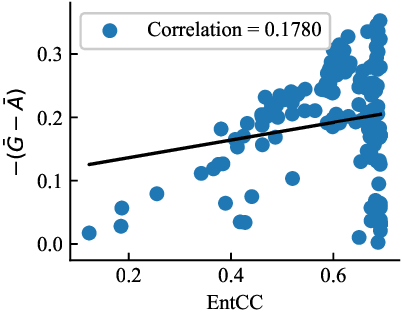}}
\subfloat[]{\centering
    \includegraphics[scale=0.4124]{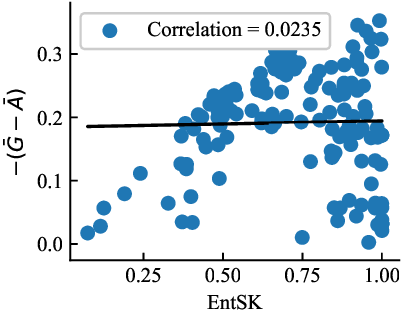}}
\subfloat[]{\centering
    \includegraphics[scale=0.4124]{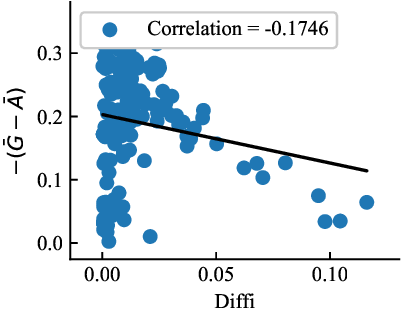}}
\subfloat[]{\centering
    \includegraphics[scale=0.4124]{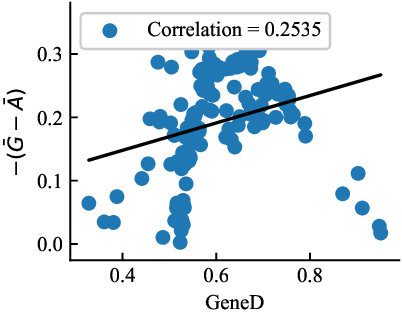}}
\subfloat[]{\centering
    \includegraphics[scale=0.4124]{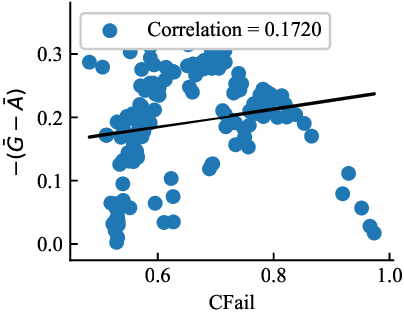}}
\caption{%
Relationship of the loss difference $-(\bar{G}-\bar{A})$ in Eq.~\eqref{eq:5,th1} and different diversity measures, using AdaBoost to conduct ensemble classifiers and MLPs as individual classifiers. 
Diversity measures used in (a--l) refer to those in Figure~\ref{fig:decom:Bag,NN}.
}\label{fig:decom:Ada,NN}
\end{minipage}
\end{figure*}

The inputs of this method are: 
(i) a training set $S$ of valid data instances, 
(ii) a set $F$ of trained individual classifiers, \ie{} the original set of classifiers to constitute an ensemble classifier in Eq.~\eqref{eq:1}, 
and (iii) the ratio $\alpha$ which means the individual classifiers in the pruned sub-ensemble are chosen in a ratio of $\alpha(\%)$ from the original ensemble. 
Notice that all individual members are treated equally here, which means that $w_i=1/|F|,\,\forall\,i\in[|F|]$. 
\begin{algorithm}[tbhp]
\centering\small
\caption{\small \fdfull{} (\fdabbr{})}
\label{framework}
\begin{algorithmic}[1]
	\REQUIRE %
	A training set $S= \{(\mathbf{x}_i,y_i) \,|\, i\in[|S|]$, 
	an original ensemble $F= \{f_j \,|\, j\in[|F|]$, 
	an arbitrary diversity measure $DIV$, 
	and the threshold $\alpha$.
	\ENSURE The pruned sub-ensemble $H \,(H\subset F)$.
	\STATE $H = \emptyset$. 
	\REPEAT
	    \STATE Compute the diversity of the ensemble on each data instance using the specified diversity measure $DIV$, \ie{} $DIV(\mathbf{x})$. \label{alg:b6}
	    \STATE Compute the margin of the ensemble on each data instance, \ie{} $\lambda-2 DIV(\mathbf{x})$, where $\lambda$ is defined in Eq.~\eqref{eq:13,lambda}. \label{alg:b7} 
	    \STATE Search for the specific data instance $(\mathbf{x},y)$, \ie{} $\argmin_{(\mathbf{x},y)\in S} \lambda-2 DIV(\mathbf{x})$. \label{alg:b3}
		\STATE Sort classifiers in $F$ that classify this instance correctly in ascending order according to the accuracy performance. \label{alg:b4}
		\STATE Move the top one $f(\cdot)$ in the previous step from $F$ to $H$\,. \label{alg:b5}
	\UNTIL{The termination condition is satisfied. }
\end{algorithmic}
\end{algorithm}
For example, individual members in one ensemble generated by AdaBoost might have different values of weights in the combination, however, their weight coefficients would still be treated as important as each other in the pruning process, 
to simplify the analyses since it doesn't talk more about the situation where the individual classifiers in the ensemble are not treated equally. 
The output of this method is the set of classifiers $H$ composing the pruned sub-ensemble after pruning, which is initially set to $\emptyset$. 
This algorithm terminates when the number of $H$ reaches the number of a pruned sub-ensemble classifier (\ie{} $\alpha|F|$), or when it cannot pick up an individual classifier in $F$ that satisfies the corresponding condition (\ie{} all individual classifiers in $F$ misclassify the corresponding instance).

The objective of this method is to obtain a pruned sub-ensemble that approaches the performance as optimal as possible. 
As shown in Algorithm~\ref{algorithm}, it firstly picks out an individual classifier considering diversity and accuracy simultaneously, raising diversity in line~\ref{alg:3} and increasing accuracy in line~\ref{alg:4}. 
Specifically, the sorting principle we use in line~\ref{alg:4} is mainly based on the accuracy performance of individual classifiers. 
Then it puts the selected individual classifier into the candidate classifier set $H$, as shown in line~\ref{alg:5}. 
Subsequently, it follows a cycle to select the expected classifiers in $F$ repeatedly until the termination conditions are met.

\begin{figure}[tbhp]
\centering %
\subfloat[]{\centering\label{fig:decom:Diver,NN,Bag}
    \includegraphics[scale=0.434]{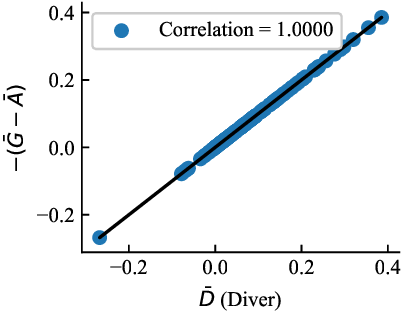}}
\hspace{2.5em}
\subfloat[]{\centering\label{fig:decom:Diver,NN,Ada}
    \includegraphics[scale=0.434]{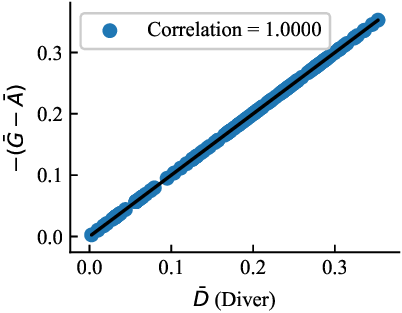}}
\caption{%
Relationship of the loss difference $-(\bar{G}-\bar{A})$ in Eq.~\eqref{eq:5,th1} and the proposed diversity measure in Eq.~\eqref{eq:6,th1}, using MLPs as individual classifiers. 
(a) Using Bagging to conduct ensembles. 
(b) Using AdaBoost to conduct ensembles.
}\label{fig:decom:Diver,NN}
\end{figure}

Moreover, inspired by the proposed ensemble pruning method (\pdabbr), we propose an ensemble pruning framework by utilizing the trade-off between accuracy and diversity, named as ``\emph{\fdfull{} (}\fdabbr\emph{)},'' presented in Algorithm~\ref{framework}.
The diversity measure used in \fdabbr{} could be any existing diversity measure, such as $Q$-statistic or the disagreement measure.
Note that \pdabbr{} is one particular case of \fdabbr{} by using the proposed diversity measure in this paper.

\begin{figure}[t]
\centering%
\subfloat[]{\label{fig:eg1:B-NB-partset:a}\centering
	\includegraphics[scale=0.614]{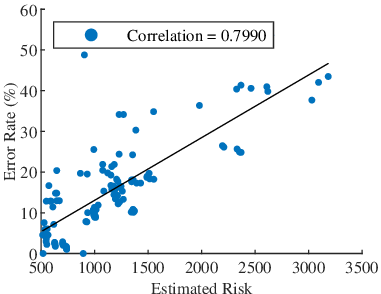}}%
\hspace{0.3ex}%
\subfloat[]{\label{fig:eg1:B-LM2-partset:b}\centering
	\includegraphics[scale=0.614]{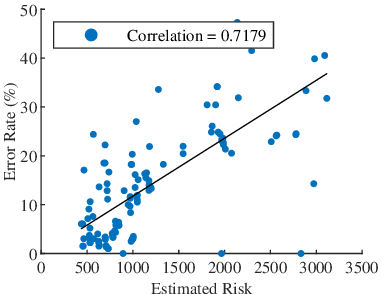}}
\caption{%
Relationship of error rate and estimated risk in Eq.~\eqref{eq:25} calculated based on diversity in Eq.~\eqref{eq:11} using bagging for binary classification. 
(a) Using NBs as individual classifiers. 
(b) Using LMs as individual classifiers.
}\label{fig:eg1:B-partset}%
\end{figure}

\subsection{Complexity Analysis of \pdabbr}
\label{prop:EPBD,complexity}

In this subsection, we give the complexity analysis of the proposed \pdabbr{}. 
According to the Algorithm~\ref{algorithm}, the computational complexity of \pdabbr{} is analyzed as follows: 
\begin{itemize}
    \item Firstly, the complexity of line~\ref{alg:3} is $\mathcal{O}(jm+m\log{}m)$ where $j=n-i+1$, and $n,m$ are the number of individual classifiers and instances, respectively, when $i\in\{1,...,k\}$. Note that $k$ is the size of the pruned sub-ensemble. 
    \item Secondly, the complexity of line~\ref{alg:4} is $\mathcal{O}(j\log{}j)$ where $j=n-i+1$, when $i\in\{1,...,k\}$. 
    \item Thirdly, the complexity of line~\ref{alg:5} is $\mathcal{O}(1)$. 
\end{itemize}
Therefore, the overall computational complexity of \pdabbr{} is $\mathcal{O}\big(\sum_{j=n-k+1}^n (jm+m\log{}m+j\log{}j)\big)$, \ie{} $\mathcal{O}\big(km(n-\frac{k-1}{2})+km\log{}m +\sum_{i=n-k+1}^n (i\log{}i)\big)$.

\begin{figure}[t]
\centering %
\begin{minipage}{0.94\linewidth}
\centering
\subfloat[]{\label{fig:eg3:BagDT:a}\centering
    \includegraphics[scale=0.575]{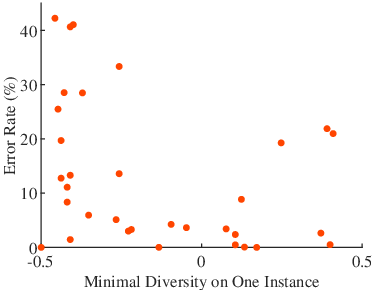}}
\hspace{0.5em}
\subfloat[]{\label{fig:eg3:BagDT:b}\centering
    \includegraphics[scale=0.575]{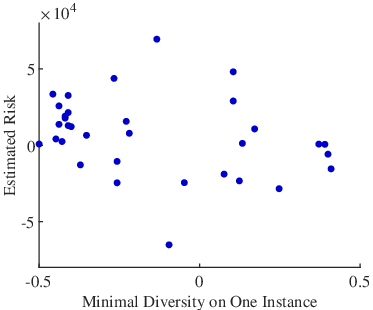}}
\caption{%
Relationship of diversity and ensemble performance in Theorem~\ref{th2}, using bagging with DTs as individual classifiers for binary classification.  
(a) Relationship between the proposed diversity and error rate of the ensemble. 
(b) Relationship between the proposed diversity and the estimated risk in Eq.~\eqref{eq:25}.
}\label{fig:eg3:BagDT}
\end{minipage}
\hspace{0.3em}
\begin{minipage}{0.94\linewidth}
\centering
\subfloat[]{\label{fig:eg3:AdaLM1:a}\centering
    \includegraphics[scale=0.575]{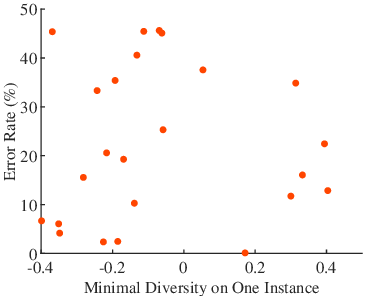}}
\hspace{0.5em}
\subfloat[]{\label{fig:eg3:AdaLM1:b}\centering
    \includegraphics[scale=0.575]{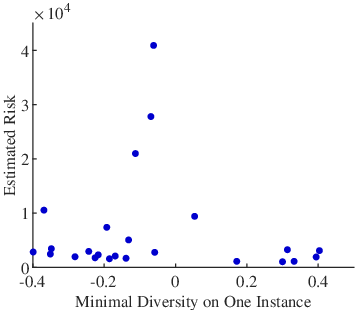}}
\caption{
Relationship of diversity and ensemble performance in Theorem~\ref{th2}, using AdaBoost with LMs as individual classifiers for binary classification.  
(a) Relationship between the proposed diversity and error rate of the ensemble. 
(b) Relationship between the proposed diversity and the estimated risk in Eq.~\eqref{eq:25}. 
}\label{fig:eg3:AdaLM1}
\end{minipage}
\end{figure}

\begin{table*}[th]
\centering
\caption{
Comparison of the state-of-the-art pruning methods with \pdabbr{} using bagging to produce homogeneous ensembles. 
The column named ``Ensem'' is the corresponding result for the entire ensemble without pruning. 
The best accuracy with a lower standard deviation is indicated with bold fonts for each data set (row).  
}\label{tab,pru:bagging}
\renewcommand\tabcolsep{1.2pt}
\scalebox{0.644}{%
\begin{threeparttable}[b]
\begin{tabular}{lc cccc ccc ccccc cc c}
	\toprule
	Data Set & Ensem & ES & KL & KP & OO & DREP & SEP & PEP & MRMR & MRMC & DiscEP & TSP.AP & TSP.DP & GMA & LCS & \pdabbr{} \\ 
	\midrule
waveform    &	89.09$\pm$1.27 &	84.34$\pm$8.76 &	83.48$\pm$8.39 &	83.48$\pm$8.39 &	89.47$\pm$0.40 &	89.35$\pm$0.66 &	89.03$\pm$0.83 &	89.19$\pm$0.81 &	88.43$\pm$0.94$\ddagger$ &	89.43$\pm$0.82 &	89.37$\pm$0.38 &	\textbf{89.71$\pm$0.75} &	88.19$\pm$0.95$\ddagger$ &	88.51$\pm$1.29 &	88.95$\pm$1.37 &	89.53$\pm$0.48 \\
heart       &	55.56$\pm$0.00 &	55.56$\pm$0.00 &	55.56$\pm$0.00 &	55.56$\pm$0.00 &	55.56$\pm$0.00 &	55.56$\pm$0.00 &	55.56$\pm$0.00 &	\textbf{55.93$\pm$0.74} &	\textbf{55.93$\pm$0.74} &	\textbf{55.93$\pm$0.74} &	\textbf{55.93$\pm$0.74} &	55.56$\pm$0.00 &	55.56$\pm$0.00 &	55.56$\pm$0.00 &	55.56$\pm$0.00 &	\textbf{55.93$\pm$0.74} \\
page        &	91.26$\pm$0.22$\ddagger$ &	91.28$\pm$0.19 &	91.26$\pm$0.24 &	91.33$\pm$0.29 &	91.37$\pm$0.29 &	91.35$\pm$0.20 &	91.24$\pm$0.11 &	91.28$\pm$0.27 &	91.33$\pm$0.20 &	91.19$\pm$0.20$\ddagger$ &	91.24$\pm$0.16 &	91.26$\pm$0.19 &	\textbf{91.39$\pm$0.21} &	91.12$\pm$0.16$\ddagger$ &	91.12$\pm$0.16$\ddagger$ &	91.39$\pm$0.26 \\
sensor (2d) &	94.12$\pm$0.80$\ddagger$ &	94.08$\pm$0.95$\ddagger$ &	94.15$\pm$0.98 &	93.97$\pm$0.93$\ddagger$ &	94.21$\pm$1.08 &	94.37$\pm$1.01 &	94.02$\pm$0.92$\ddagger$ &	\textbf{94.41$\pm$0.83} &	94.24$\pm$0.55 &	94.21$\pm$1.01 &	93.73$\pm$0.95$\ddagger$ &	94.10$\pm$1.09 &	94.08$\pm$0.80 &	94.02$\pm$1.04$\ddagger$ &	94.10$\pm$1.02 &	94.37$\pm$0.85 \\
sensor (24d)&	89.37$\pm$0.61 &	89.13$\pm$0.27 &	88.95$\pm$0.73 &	88.91$\pm$0.75 &	\textbf{89.39$\pm$1.38} &	88.82$\pm$1.05$\ddagger$ &	88.82$\pm$1.11$\ddagger$ &	89.31$\pm$1.25 &	89.15$\pm$0.97 &	89.09$\pm$0.92 &	88.87$\pm$0.86 &	88.93$\pm$0.93$\ddagger$ &	88.95$\pm$1.02 &	89.07$\pm$0.93 &	88.74$\pm$1.16 &	89.33$\pm$0.95 \\
EEGEyeState &	55.13$\pm$0.01 &	\textbf{55.13$\pm$0.00} &	\textbf{55.13$\pm$0.00} &	\textbf{55.13$\pm$0.00} &	55.13$\pm$0.01 &	\textbf{55.13$\pm$0.00} &	\textbf{55.13$\pm$0.00} &	\textbf{55.13$\pm$0.00} &	55.13$\pm$0.01 &	55.13$\pm$0.01 &	55.13$\pm$0.01 &	\textbf{55.13$\pm$0.00} &	55.13$\pm$0.01 &	\textbf{55.13$\pm$0.00} &	\textbf{55.13$\pm$0.00} &	55.13$\pm$0.01 \\
gmm (10d)   &	99.70$\pm$0.60 &	99.70$\pm$0.60 &	99.70$\pm$0.60 &	99.70$\pm$0.60 &	\textbf{99.80$\pm$0.40} &	99.70$\pm$0.60 &	99.70$\pm$0.60 &	\textbf{99.80$\pm$0.40} &	\textbf{99.80$\pm$0.40} &	\textbf{99.80$\pm$0.40} &	\textbf{99.80$\pm$0.40} &	99.70$\pm$0.60 &	99.60$\pm$0.59 &	99.70$\pm$0.60 &	99.60$\pm$0.59 &	\textbf{99.80$\pm$0.40} \\
ames        &	61.17$\pm$4.01 &	58.54$\pm$7.61 &	55.77$\pm$6.67$\ddagger$ &	56.50$\pm$4.72 &	59.71$\pm$6.93 &	57.96$\pm$5.57$\ddagger$ &	58.54$\pm$4.53 &	59.85$\pm$5.28 &	59.12$\pm$4.21 &	57.66$\pm$3.72 &	\textbf{61.61$\pm$4.49} &	60.44$\pm$4.78 &	59.71$\pm$6.04 &	56.50$\pm$2.64 &	59.56$\pm$3.53 &	61.02$\pm$5.80 \\
wisconsin   &	96.89$\pm$0.55 &	96.44$\pm$0.55 &	96.44$\pm$0.86 &	96.89$\pm$0.73 &	96.59$\pm$1.00 &	96.59$\pm$0.76 &	96.00$\pm$0.59 &	96.44$\pm$0.98 &	96.74$\pm$0.76 &	96.89$\pm$1.09 &	97.04$\pm$0.47 &	96.74$\pm$1.00 &	96.89$\pm$0.55 &	96.59$\pm$0.36 &	\textbf{96.89$\pm$0.30} &	96.89$\pm$0.73 \\
ecoli       &	95.76$\pm$2.23 &	95.76$\pm$2.42 &	95.76$\pm$2.42 &	95.76$\pm$2.42 &	95.76$\pm$2.42 &	95.76$\pm$2.42 &	95.76$\pm$2.42 &	\textbf{96.36$\pm$2.06} &	\textbf{96.36$\pm$2.06} &	95.76$\pm$2.23 &	95.76$\pm$2.23 &	95.76$\pm$2.23 &	95.76$\pm$2.42 &	96.06$\pm$2.06 &	96.06$\pm$2.27 &	\textbf{96.36$\pm$2.06} \\
liver       &	62.90$\pm$2.98$\ddagger$ &	61.16$\pm$2.49 &	62.03$\pm$5.75 &	65.22$\pm$2.43 &	63.77$\pm$3.04 &	65.51$\pm$3.60 &	64.64$\pm$3.38 &	66.38$\pm$2.96 &	65.80$\pm$2.98 &	63.19$\pm$2.98 &	62.90$\pm$2.84 &	61.74$\pm$2.98 &	64.93$\pm$3.82 &	61.45$\pm$6.12 &	61.74$\pm$4.36$\ddagger$ &	\textbf{66.96$\pm$3.36} \\
yeast       &	69.59$\pm$4.61 &	67.77$\pm$2.86$\ddagger$ &	67.70$\pm$2.36$\ddagger$ &	69.39$\pm$1.66 &	69.73$\pm$3.27 &	65.41$\pm$2.60$\ddagger$ &	69.80$\pm$4.23 &	69.39$\pm$3.44 &	69.86$\pm$3.06 &	69.12$\pm$2.89 &	68.24$\pm$3.82$\ddagger$ &	68.58$\pm$3.15 &	68.72$\pm$3.44 &	67.36$\pm$3.66$\ddagger$ &	68.45$\pm$3.95 &	\textbf{70.54$\pm$2.92} \\
sonar       &	80.00$\pm$6.62 &	76.59$\pm$5.69 &	78.05$\pm$6.90 &	79.02$\pm$6.28 &	80.49$\pm$6.54 &	80.00$\pm$5.85 &	79.51$\pm$3.65 &	79.51$\pm$5.89 &	\textbf{81.95$\pm$8.10} &	80.49$\pm$6.36 &	76.10$\pm$4.97 &	79.02$\pm$3.31 &	76.10$\pm$6.79 &	79.51$\pm$7.49 &	80.49$\pm$6.17 &	80.49$\pm$7.07 \\
wilt        &	93.90$\pm$1.34 &	93.71$\pm$1.49 &	94.37$\pm$0.85 &	93.34$\pm$1.51 &	94.33$\pm$0.87 &	92.12$\pm$1.81$\ddagger$ &	93.90$\pm$1.44 &	93.59$\pm$1.59 &	94.33$\pm$0.98 &	94.35$\pm$0.93 &	94.00$\pm$1.25 &	94.46$\pm$0.95 &	93.86$\pm$1.00 &	93.92$\pm$1.44 &	93.75$\pm$1.17 &	\textbf{94.48$\pm$0.87} \\
spam        &	78.67$\pm$3.44 &	78.54$\pm$2.74 &	77.34$\pm$2.18 &	73.32$\pm$5.74 &	74.56$\pm$6.96 &	79.09$\pm$2.84 &	77.87$\pm$1.39 &	79.61$\pm$2.67 &	79.04$\pm$2.80 &	72.82$\pm$8.46 &	78.91$\pm$3.13 &	79.76$\pm$2.42 &	77.87$\pm$3.55 &	73.93$\pm$5.00 &	77.39$\pm$1.58 &	\textbf{79.80$\pm$2.41} \\
landsat     &	96.25$\pm$0.63 &	96.31$\pm$0.63 &	95.57$\pm$0.98 &	95.30$\pm$1.02 &	96.27$\pm$0.58 &	95.68$\pm$0.84 &	96.31$\pm$0.54 &	95.96$\pm$0.91 &	96.22$\pm$0.56 &	95.65$\pm$1.16 &	96.22$\pm$0.61 &	\textbf{96.33$\pm$0.50} &	95.69$\pm$0.90 &	96.13$\pm$0.75 &	96.07$\pm$0.58 &	96.27$\pm$0.65 \\
	\hline
	$t$-test (W/T/L) & 3/13/0 & 2/14/0 & 2/14/0 & 1/15/0 & 0/16/0 & 4/12/0 & 2/14/0 & 0/16/0 & 1/15/0 & 1/15/0 & 2/14/0 & 1/15/0 & 1/15/0 & 4/12/0 & 2/14/0 & --- \\
	Average Rank & 7.47 & 10.84 & 11.47 & 11.03 & 6.00 & 9.41 & 9.78 & 6.56 & 5.13 & 7.88 & 8.28 & 8.22 & 9.28 & 11.44 & 10.69 & 2.53 \\
	\bottomrule
\end{tabular}
\begin{tablenotes}
\item[1] The reported results are the average test accuracy (\%) of each method and the corresponding standard deviation under 5-fold cross-validation on each data set. 
\item[2] By two-tailed paired $t$-test at 5\% significance level, $\ddagger$ and $\dagger$ denote that the performance of \pdabbr{} is superior to and inferior to that of the comparative method, respectively. 
\item[3] The last two rows show the results of $t$-test and average rank, respectively. 
The ``W/T/L'' in $t$-test indicates that \pdabbr{} is superior to, not significantly different from, or inferior to the corresponding comparative methods. The average rank is calculated according to the Friedman test~\cite{demvsar2006statistical}. 
\end{tablenotes}
\end{threeparttable}
}
\end{table*}

\begin{table*}[th]
\centering
\caption{
Comparison of the state-of-the-art pruning methods with \pdabbr{} using AdaBoost to produce homogeneous ensembles. 
The column named ``Ensem'' is the corresponding result for the entire ensemble without pruning. 
The best accuracy with a lower standard deviation is indicated with bold fonts for each data set (row).  
}\label{tab,pru:adaboost}
\renewcommand\tabcolsep{1.2pt}
\scalebox{0.644}{%
\begin{threeparttable}[b]
\begin{tabular}{lc cccc ccc ccccc cc c}
	\toprule
	Data Set & Ensem & ES & KL & KP & OO & DREP & SEP & PEP & MRMR & MRMC & DiscEP & TSP.AP & TSP.DP & GMA & LCS & \pdabbr{} \\ 
	\midrule
spam        &	79.33$\pm$2.04 &	81.44$\pm$1.58 &	79.39$\pm$3.15 &	78.32$\pm$2.76 &	79.17$\pm$2.23 &	79.91$\pm$1.44 &	79.06$\pm$2.90 &	\textbf{81.85$\pm$1.36} &	81.44$\pm$1.58 &	78.78$\pm$2.24 &	78.78$\pm$2.24 &	79.04$\pm$2.11 &	76.63$\pm$2.51 &	73.54$\pm$6.13 &	79.04$\pm$2.11 &	79.98$\pm$2.12 \\
credit      &	77.96$\pm$0.06 &	77.93$\pm$0.06 &	77.93$\pm$0.13 &	77.91$\pm$0.12 &	77.99$\pm$0.08 &	77.99$\pm$0.09 &	77.93$\pm$0.09 &	77.95$\pm$0.08 &	77.98$\pm$0.14 &	77.98$\pm$0.09 &	77.96$\pm$0.14 &	77.91$\pm$0.13 &	77.88$\pm$0.11$\ddagger$ &	77.86$\pm$0.10 &	77.95$\pm$0.06 &	\textbf{78.03$\pm$0.10} \\
page        &	90.40$\pm$0.12$\ddagger$ &	90.35$\pm$0.26$\ddagger$ &	90.40$\pm$0.27$\ddagger$ &	90.29$\pm$0.33$\ddagger$ &	90.93$\pm$0.32 &	\textbf{90.99$\pm$0.24} &	90.49$\pm$0.32 &	90.88$\pm$0.35 &	90.60$\pm$0.25 &	90.93$\pm$0.32 &	90.91$\pm$0.31 &	\textbf{90.99$\pm$0.24} &	90.22$\pm$0.42$\ddagger$ &	90.80$\pm$0.32 &	90.77$\pm$0.38 &	\textbf{90.99$\pm$0.24} \\
shuttle     &	96.75$\pm$0.08$\ddagger$ &	97.33$\pm$0.06$\ddagger$ &	97.33$\pm$0.06$\ddagger$ &	92.40$\pm$2.09$\ddagger$ &	97.81$\pm$0.10 &	97.70$\pm$0.05$\ddagger$ &	94.75$\pm$2.43 &	96.80$\pm$2.10 &	\textbf{97.85$\pm$0.05} &	97.65$\pm$0.10$\ddagger$ &	\textbf{97.85$\pm$0.05} &	\textbf{97.85$\pm$0.05} &	92.05$\pm$2.09$\ddagger$ &	97.59$\pm$0.12$\ddagger$ &	\textbf{97.85$\pm$0.05} &	\textbf{97.85$\pm$0.05} \\
wilt        &	94.64$\pm$0.04 &	94.66$\pm$0.08 &	94.66$\pm$0.08 &	94.66$\pm$0.08 &	94.66$\pm$0.08 &	94.66$\pm$0.08 &	94.62$\pm$0.00 &	94.62$\pm$0.00 &	\textbf{94.71$\pm$0.10} &	94.66$\pm$0.08 &	\textbf{94.71$\pm$0.10} &	\textbf{94.71$\pm$0.10} &	\textbf{94.71$\pm$0.10} &	94.66$\pm$0.08 &	\textbf{94.71$\pm$0.10} &	\textbf{94.71$\pm$0.10} \\
segmentation&	\textbf{67.40$\pm$0.56} &	65.93$\pm$0.67$\ddagger$ &	65.19$\pm$1.24$\ddagger$ &	63.94$\pm$0.95$\ddagger$ &	67.14$\pm$1.14 &	68.14$\pm$1.07$\dagger$ &	66.02$\pm$1.79 &	66.32$\pm$3.21 &	65.89$\pm$0.87$\ddagger$ &	65.71$\pm$0.95$\ddagger$ &	66.67$\pm$0.43 &	66.28$\pm$0.73 &	65.50$\pm$1.66 &	66.15$\pm$0.67 &	66.54$\pm$0.45 &	66.80$\pm$0.56 \\
iono        &	81.43$\pm$1.56 &	77.71$\pm$5.16 &	78.00$\pm$7.15 &	80.57$\pm$5.16 &	\textbf{84.29$\pm$3.61} &	83.71$\pm$3.68 &	80.57$\pm$6.43 &	81.43$\pm$6.32 &	82.57$\pm$3.43 &	82.00$\pm$5.90 &	82.57$\pm$2.77 &	82.57$\pm$2.77 &	76.00$\pm$6.29 &	72.86$\pm$4.52$\ddagger$ &	80.57$\pm$5.16 &	84.00$\pm$2.91 \\
wilt        &	97.17$\pm$0.24 &	96.38$\pm$0.52 &	97.35$\pm$0.38 &	95.37$\pm$1.25$\ddagger$ &	97.29$\pm$0.65 &	97.35$\pm$0.38 &	97.13$\pm$0.56 &	97.31$\pm$0.44 &	97.27$\pm$0.29 &	97.25$\pm$0.51 &	\textbf{97.37$\pm$0.40} &	97.35$\pm$0.38 &	96.34$\pm$1.28 &	97.35$\pm$0.38 &	97.13$\pm$0.48 &	97.35$\pm$0.38 \\
ecoli       &	\textbf{69.70$\pm$2.71} &	64.24$\pm$4.35 &	62.73$\pm$4.94 &	63.33$\pm$3.76 &	66.36$\pm$6.24 &	64.55$\pm$3.26 &	63.94$\pm$3.76 &	63.03$\pm$4.75 &	64.55$\pm$3.12 &	61.82$\pm$4.64 &	60.91$\pm$8.10 &	66.06$\pm$6.82 &	65.15$\pm$1.66 &	64.55$\pm$4.94 &	62.73$\pm$4.66 &	67.27$\pm$4.85 \\
yeast       &	63.45$\pm$1.90 &	\textbf{66.15$\pm$2.35} &	63.24$\pm$1.94 &	63.31$\pm$1.04 &	63.78$\pm$2.80 &	60.61$\pm$2.65 &	61.96$\pm$1.46 &	63.18$\pm$2.73 &	62.84$\pm$3.64 &	62.64$\pm$3.29 &	62.50$\pm$2.87 &	63.18$\pm$2.79 &	59.80$\pm$3.74 &	62.09$\pm$2.85 &	62.50$\pm$2.45 &	63.78$\pm$3.60 \\
ringnorm    &	68.32$\pm$1.27$\ddagger$ &	68.67$\pm$1.21 &	68.21$\pm$1.69 &	67.88$\pm$0.68$\ddagger$ &	69.20$\pm$0.90 &	\textbf{69.86$\pm$1.31} &	67.05$\pm$1.40$\ddagger$ &	68.55$\pm$1.22 &	69.17$\pm$0.82 &	68.53$\pm$1.03 &	68.88$\pm$1.10 &	69.32$\pm$1.04 &	68.15$\pm$1.22$\ddagger$ &	68.80$\pm$0.86 &	68.86$\pm$0.77 &	69.17$\pm$0.92 \\
waveform    &	83.46$\pm$0.60 &	83.00$\pm$0.64 &	82.52$\pm$0.62 &	81.72$\pm$1.11$\ddagger$ &	82.78$\pm$1.21 &	83.26$\pm$0.78 &	82.82$\pm$1.03$\ddagger$ &	82.86$\pm$0.74 &	83.44$\pm$0.19 &	83.14$\pm$1.14 &	82.40$\pm$0.94$\ddagger$ &	82.90$\pm$0.30$\ddagger$ &	82.80$\pm$1.05 &	82.32$\pm$0.67$\ddagger$ &	82.54$\pm$0.47$\ddagger$ &	\textbf{83.86$\pm$0.74} \\
gmm (10d)   &	99.70$\pm$0.25 &	\textbf{99.80$\pm$0.25} &	95.58$\pm$5.24 &	97.59$\pm$4.58 &	99.70$\pm$0.25 &	99.60$\pm$0.38 &	99.70$\pm$0.60 &	\textbf{99.80$\pm$0.25} &	97.39$\pm$4.48 &	99.60$\pm$0.38 &	97.49$\pm$4.53 &	99.50$\pm$0.32 &	99.50$\pm$0.32 &	99.70$\pm$0.25 &	99.60$\pm$0.38 &	\textbf{99.80$\pm$0.25} \\
	\hline
	$t$-test (W/T/L) & 3/10/0 & 3/10/0 & 3/10/0 & 6/7/0 & 0/13/0 & 1/11/1 & 2/11/0 & 0/13/0 & 1/12/0 & 2/11/0 & 1/12/0 & 1/12/0 & 4/9/0 & 3/10/0 & 1/12/0 & --- \\
	Average Rank & 7.62 & 8.85 & 10.96 & 13.15 & 5.35 & 5.42 & 11.19 & 8.08 & 6.58 & 9.23 & 8.04 & 6.35 & 12.77 & 10.58 & 9.15 & 2.69 \\
	\bottomrule
\end{tabular}
\end{threeparttable}
}
\end{table*}

\begin{figure*}[tbhp]
\centering
\begin{minipage}{\linewidth}
\centering %
\subfloat[]{\label{fig,expt4sbi:bag,a}
    \includegraphics[scale=0.5461]{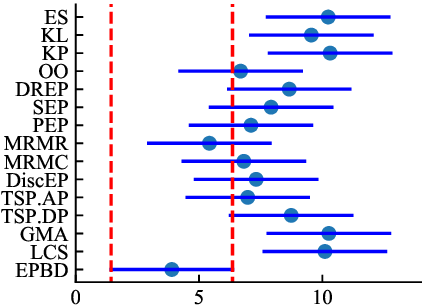}}
\hspace{0.5em}
\subfloat[]{\label{fig,expt4sbi:bag,b}
    \includegraphics[scale=0.5461]{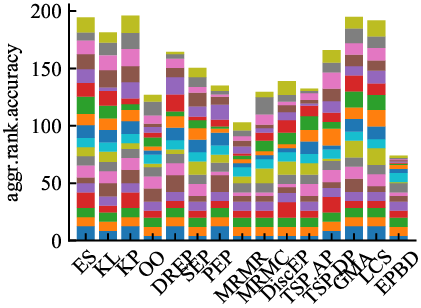}}
\hspace{0.5em}
\subfloat[]{\label{fig,expt4sbi:bag,c}
    \includegraphics[scale=0.5461]{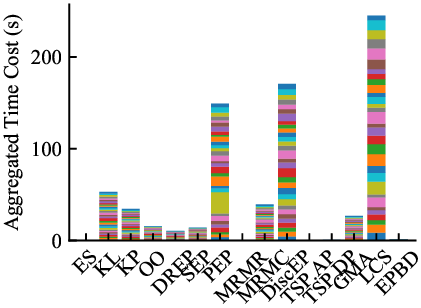}}
\hspace{0.5em}
\subfloat[]{\label{fig,expt4sbi:bag,d}
    \includegraphics[scale=0.5461]{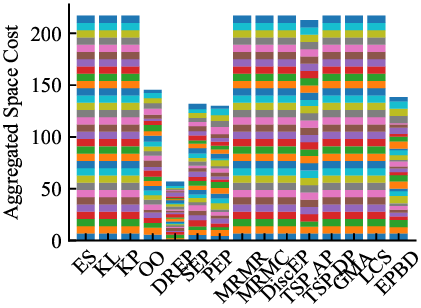}}
\caption{
Comparison of the state-of-the-art methods with \pdabbr{} on the test accuracy, using bagging to produce homogeneous ensembles. 
(a) Friedman test chart (non-overlapping means significant difference) \citep{demvsar2006statistical}. 
(b) The aggregated rank for each pruning method (the smaller the better) \citep{qian2015pareto}. 
(c) The aggregated time cost for each method. 
(d) The aggregated space cost for each method. 
}\label{fig,expt4s:Bag,cost,NN}
\vspace{0.8em}
\end{minipage}
\begin{minipage}{\linewidth}
\centering
\subfloat[]{
    \includegraphics[scale=0.5461]{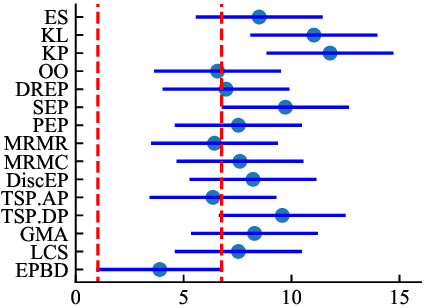}}
\hspace{0.5em}
\subfloat[]{
    \includegraphics[scale=0.5461]{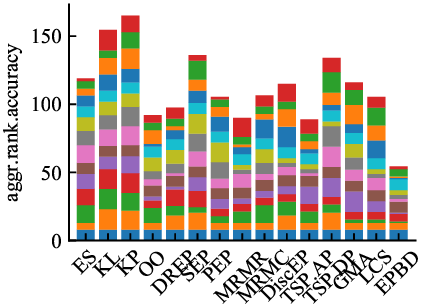}}
\hspace{0.5em}
\subfloat[]{\label{fig,expt4sbi:ada,c}
    \includegraphics[scale=0.5461]{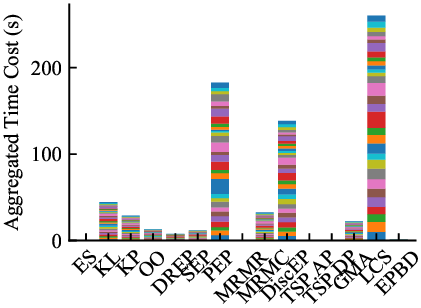}}
\hspace{0.5em}
\subfloat[]{\label{fig,expt4sbi:ada,d}
    \includegraphics[scale=0.5461]{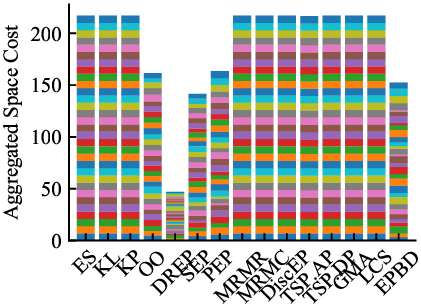}}
\caption{
Comparison of the state-of-the-art methods with \pdabbr{} on the test accuracy, using AdaBoost to produce homogeneous ensembles. 
(a) Friedman test chart (non-overlapping means significant difference) \citep{demvsar2006statistical}. 
(b) The aggregated rank for each pruning method (the smaller the better) \citep{qian2015pareto}. 
(c) The aggregated time cost for each method. 
(d) The aggregated space cost for each method. 
}\label{fig,expt4s:Ada,cost,KNNd}
\end{minipage}
\end{figure*}

\section{Experiments}
\label{experiment}

In this section, we elaborate on our experiments to evaluate the proposed relationship between the proposed diversity and ensemble generalization performance and the proposed algorithms.  
The data sets we use include one image data set with $12,500$ pictures (Dogs vs Cats\footnote{\url{http://www.kaggle.com/c/dogs-vs-cats}}) and $28$ data sets from UCI repository\footnote{\url{http://archive.ics.uci.edu/ml/datasets.html}}~\cite{Lichman:2013}. 
Standard $5$-fold cross-validation is used in these experiments, \ie{} in each iteration, the entire data set is split into two parts, 
with $80\%$ as the training set and $20\%$ as the test set. 
Bagging~\cite{breiman1996bagging} and AdaBoost~\cite{freund1995boosting,freund1996experiments} are used to constitute an ensemble classifier on various kinds of individual classifiers including decision trees (DT), naive Bayesian (NB) classifiers, $k$-nearest neighbors (KNN) classifiers, linear model (LM) classifiers, linear SVMs (LSVM), support vector machines (SVM), and multilayer perceptrons (MLP). 
We evaluate the proposed relationship between the proposed diversity and ensemble performance using scatter diagrams with their correlation. 
Besides, to evaluate our proposed pruning method \pdabbr{}, the baselines we considered are a variety of ranking-based methods, namely Early Stopping (ES), KL-divergence pruning (KL), Kappa pruning (KP)~\cite{margineantu1997pruning}, orientation ordering pruning (OO)~\cite{martinez2006pruning}, diversity regularized ensemble pruning (DREP)~\cite{li2012diversity}, accuracy-based ensemble pruning (TSP.AP) \citep{zhang2019two}, and diversity-based ensemble pruning (TSP.DP) \citep{zhang2019two} as well as optimization-based methods, namely single-objective ensemble pruning (SEP), and Pareto ensemble pruning (PEP)~\cite{qian2015pareto}, the maximal relevant minimal redundant (MRMR) method \citep{xia2018maximum}, the maximal relevancy miximum complementary (MRMC) method \citep{xia2018maximum}, and the discriminative ensemble pruning (DiscEP) \citep{cao2018optimizing}. 
Moreover, we adopt two methods from diversity maximization via composable coresets~\cite{aghamolaei2015diversity,indyk2014composable} and change them slightly to make them suitable for pruning problems, namely the Gonzalez's algorithm (GMA) and local search algorithm (LCS). 
An ensemble is trained and pruned on the training set and then evaluated on the test set. 
It is worth mentioning that several methods (such as OO, DREP, SEP, PEP, LCS, and TSP.AP) cannot fix the number of classifiers after ensemble pruning, while others could do that by giving a pruning rate that is the up limit of the percentage of those discarded individual classifiers in the original ensemble. 
Those methods that cannot fix the size of the pruned sub-ensemble might lead to increasing or reducing the size of pruned sub-ensembles and affect their space cost.

\begin{table*}[th]
\centering
\caption{
Comparison of different diversity measures with \fdabbr{} using bagging to produce an ensemble with NBs as individual classifiers. The column named ``Ensem'' is the corresponding result for the entire ensemble without pruning. 
The best accuracy with a lower standard deviation is indicated with bold fonts for each data set (row).  
}\label{tab,frame:bagging,NB}
\renewcommand\tabcolsep{1.8pt}
\scalebox{0.656}{
\begin{threeparttable}[b]
\begin{tabular}{lc ccccc ccccccc c}
	\toprule
	Data Set & Ensem & Disag & QStat & Corre & KStat & DoubF & KWVar & Inter & EntCC & EntSK & Diffi & GeneD & CFail & \pdabbr{} \\ 
	\midrule
ames           & 57.37$\pm$3.53$\ddagger$ &	\textbf{59.71$\pm$4.39} &	59.12$\pm$4.36 &	59.12$\pm$4.36 &	59.12$\pm$4.36 &	58.69$\pm$4.07 &	59.12$\pm$4.36 &	59.12$\pm$4.36 &	58.39$\pm$4.08 &	58.39$\pm$4.08 &	59.12$\pm$4.36 &	58.69$\pm$4.07 &	58.69$\pm$4.07 &	58.69$\pm$4.07 \\
card           & \textbf{79.56$\pm$4.48} &	79.42$\pm$5.57 &	79.12$\pm$5.03 &	79.12$\pm$5.03 &	79.12$\pm$5.03 &	79.42$\pm$5.57 &	79.42$\pm$5.57 &	79.42$\pm$5.57 &	79.56$\pm$5.44 &	79.56$\pm$5.44 &	79.12$\pm$5.03 &	79.42$\pm$5.57 &	79.42$\pm$4.85 &	79.42$\pm$5.57 \\
heart          & 84.07$\pm$3.23 &	\textbf{84.44$\pm$1.89} &	84.07$\pm$2.22 &	84.07$\pm$2.22 &	84.07$\pm$2.22 &	83.70$\pm$2.72 &	84.07$\pm$3.01 &	84.07$\pm$3.01 &	83.70$\pm$2.72 &	83.70$\pm$2.72 &	84.07$\pm$2.22 &	83.70$\pm$2.72 &	83.70$\pm$2.72 &	83.70$\pm$2.72 \\
ringnorm       & \textbf{98.69$\pm$0.24} &	98.66$\pm$0.25 &	98.66$\pm$0.25 &	98.66$\pm$0.25 &	98.66$\pm$0.25 &	98.66$\pm$0.25 &	98.66$\pm$0.25 &	98.66$\pm$0.25 &	98.66$\pm$0.25 &	98.66$\pm$0.25 &	98.66$\pm$0.25 &	98.66$\pm$0.25 &	98.66$\pm$0.25 &	98.66$\pm$0.25 \\
wisconsin      & \textbf{96.00$\pm$1.11} &	96.00$\pm$1.37 &	96.00$\pm$1.37 &	96.00$\pm$1.37 &	96.00$\pm$1.37 &	96.00$\pm$1.37 &	96.00$\pm$1.37 &	96.00$\pm$1.37 &	96.00$\pm$1.37 &	96.00$\pm$1.37 &	96.00$\pm$1.37 &	96.00$\pm$1.37 &	96.00$\pm$1.37 &	96.00$\pm$1.37 \\
landsat        & 90.30$\pm$0.73 &	\textbf{90.48$\pm$0.73} &	\textbf{90.48$\pm$0.73} &	\textbf{90.48$\pm$0.73} &	\textbf{90.48$\pm$0.73} &	\textbf{90.48$\pm$0.73} &	\textbf{90.48$\pm$0.73} &	\textbf{90.48$\pm$0.73} &	90.47$\pm$0.71 &	\textbf{90.48$\pm$0.73} &	\textbf{90.48$\pm$0.73} &	\textbf{90.48$\pm$0.73} &	90.47$\pm$0.71 &	\textbf{90.48$\pm$0.73} \\
shuttle        & 89.48$\pm$0.16$\ddagger$ &	\textbf{89.71$\pm$0.14} &	\textbf{89.71$\pm$0.14} &	\textbf{89.71$\pm$0.14} &	\textbf{89.71$\pm$0.14} &	\textbf{89.71$\pm$0.14} &	\textbf{89.71$\pm$0.14} &	\textbf{89.71$\pm$0.14} &	\textbf{89.71$\pm$0.14} &	\textbf{89.71$\pm$0.14} &	\textbf{89.71$\pm$0.14} &	\textbf{89.71$\pm$0.14} &	\textbf{89.71$\pm$0.14} &	\textbf{89.71$\pm$0.14} \\
ecoli          & 92.12$\pm$2.78 &	\textbf{92.42$\pm$3.46} &	\textbf{92.42$\pm$3.46} &	\textbf{92.42$\pm$3.46} &	\textbf{92.42$\pm$3.46} &	94.24$\pm$2.61 &	\textbf{92.42$\pm$3.46} &	\textbf{92.42$\pm$3.46} &	\textbf{92.42$\pm$3.46} &	\textbf{92.42$\pm$3.46} &	\textbf{92.42$\pm$3.46} &	\textbf{92.42$\pm$3.46} &	\textbf{92.42$\pm$3.46} &	94.24$\pm$2.61 \\
yeast          & 58.24$\pm$2.64$\ddagger$ &	60.20$\pm$1.67 &	57.30$\pm$3.33 &	57.30$\pm$3.33 &	57.30$\pm$3.33 &	\textbf{61.82$\pm$1.46} &	\textbf{61.82$\pm$1.46} &	\textbf{61.82$\pm$1.46} &	60.27$\pm$1.64 &	60.20$\pm$1.67 &	57.30$\pm$3.33 &	61.69$\pm$1.70 &	61.49$\pm$1.61 &	\textbf{61.82$\pm$1.46} \\
mammo\_graphic & \textbf{80.73$\pm$3.04} &	79.79$\pm$2.76 &	79.69$\pm$2.57 &	79.69$\pm$2.57 &	79.69$\pm$2.57 &	79.58$\pm$2.72 &	79.48$\pm$2.61 &	79.48$\pm$2.61 &	79.79$\pm$2.76 &	79.79$\pm$2.76 &	79.69$\pm$2.57 &	79.58$\pm$2.72 &	79.58$\pm$2.72 &	79.58$\pm$2.72 \\
madelon        & 58.85$\pm$2.37 &	58.96$\pm$2.28 &	59.19$\pm$1.31 &	59.19$\pm$1.31 &	59.19$\pm$1.31 &	59.04$\pm$2.07 &	59.04$\pm$2.07 &	59.04$\pm$2.07 &	\textbf{59.58$\pm$2.39} &	58.96$\pm$2.28 &	59.19$\pm$1.31 &	59.04$\pm$2.07 &	58.92$\pm$2.22 &	59.04$\pm$2.07 \\
sensor (2d)    & 81.41$\pm$0.86 &	81.61$\pm$0.91 &	81.81$\pm$0.57 &	81.81$\pm$0.57 &	81.81$\pm$0.57 &	\textbf{82.02$\pm$0.70} &	81.81$\pm$0.57 &	81.81$\pm$0.57 &	81.59$\pm$0.93 &	81.61$\pm$0.91 &	81.81$\pm$0.57 &	\textbf{82.02$\pm$0.70} &	81.92$\pm$0.76 &	\textbf{82.02$\pm$0.70} \\
sensor (4d)    & 74.57$\pm$0.88$\ddagger$ &	75.44$\pm$0.99 &	\textbf{75.55$\pm$0.89} &	\textbf{75.55$\pm$0.89} &	\textbf{75.55$\pm$0.89} &	\textbf{75.55$\pm$0.89} &	\textbf{75.55$\pm$0.89} &	\textbf{75.55$\pm$0.89} &	75.33$\pm$0.85 &	75.44$\pm$0.99 &	\textbf{75.55$\pm$0.89} &	\textbf{75.55$\pm$0.89} &	75.51$\pm$0.94 &	\textbf{75.55$\pm$0.89} \\
waveform       & 85.29$\pm$0.95 &	85.35$\pm$0.94 &	85.33$\pm$0.92 &	85.33$\pm$0.92 &	85.33$\pm$0.92 &	85.33$\pm$0.92 &	85.33$\pm$0.92 &	85.33$\pm$0.92 &	\textbf{85.35$\pm$0.91} &	85.35$\pm$0.94 &	85.33$\pm$0.92 &	85.33$\pm$0.92 &	85.33$\pm$0.92 &	85.33$\pm$0.92 \\
	\hline
	$t$-test (W/T/L) & 4/10/0 & 0/14/0 & 0/14/0 & 0/14/0 & 0/14/0 & 0/14/0 & 0/14/0 & 0/14/0 & 0/14/0 & 0/14/0 & 0/14/0 & 0/14/0 & 0/14/0 & --- \\
	Average Rank & 9.89 & 6.71 & 7.21 & 7.21 & 7.21 & 6.79 & 7.21 & 7.21 & 7.68 & 7.86 & 7.21 & 7.43 & 8.57 & 6.79 \\
	\bottomrule
\end{tabular}
\begin{tablenotes}
\item[1] The reported results are the average test accuracy (\%) of each method and the corresponding standard deviation under 5-fold cross-validation on each data set. 
\item[2] By two-tailed paired $t$-test at 5\% significance level, $\ddagger$ and $\dagger$ denote that the performance of \pdabbr{} is superior to and inferior to that of the comparative method, respectively. 
\item[3] The last two rows show the results of $t$-test and average rank, respectively. 
The ``W/T/L'' in $t$-test indicates that \pdabbr{} is superior to, not significantly different from, or inferior to the corresponding comparative methods. The average rank is calculated according to the Friedman test~\cite{demvsar2006statistical}. 
\end{tablenotes}
\end{threeparttable}
}
\end{table*}

\begin{table*}[th]
\centering
\caption{
Comparison of different diversity measures with \fdabbr{} using bagging to produce an ensemble with KNNs as individual classifiers. 
The column named ``Ensem'' is the corresponding result for the entire ensemble without pruning. 
The best accuracy with a lower standard deviation is indicated with bold fonts for each data set (row).  
}\label{tab,frame:bagging,KNNd}
\renewcommand\tabcolsep{1.8pt}
\scalebox{0.656}{%
\begin{threeparttable}[b]
\begin{tabular}{lc ccccc ccccccc c}
	\toprule
	Data Set & Ensem & Disag & QStat & Corre & KStat & DoubF & KWVar & Inter & EntCC & EntSK & Diffi & GeneD & CFail & \pdabbr{} \\ 
	\midrule
gmm (2d)       &	96.90$\pm$0.63 &	96.92$\pm$0.66 &	\textbf{96.97$\pm$0.66} &	\textbf{96.97$\pm$0.66} &	\textbf{96.97$\pm$0.66} &	\textbf{96.97$\pm$0.66} &	\textbf{96.97$\pm$0.66} &	\textbf{96.97$\pm$0.66} &	96.92$\pm$0.66 &	96.92$\pm$0.66 &	\textbf{96.97$\pm$0.66} &	\textbf{96.97$\pm$0.66} &	\textbf{96.97$\pm$0.66} &	\textbf{96.97$\pm$0.66} \\
card           &	67.74$\pm$2.33 &	68.03$\pm$2.67 &	67.74$\pm$3.28 &	67.74$\pm$3.28 &	67.74$\pm$3.28 &	\textbf{68.91$\pm$2.83} &	\textbf{68.91$\pm$2.83} &	\textbf{68.91$\pm$2.83} &	67.88$\pm$2.85 &	68.03$\pm$2.67 &	67.74$\pm$3.28 &	68.32$\pm$2.15 &	\textbf{68.91$\pm$2.83} &	\textbf{68.91$\pm$2.83} \\
liver          &	64.06$\pm$4.62 &	62.61$\pm$5.45 &	63.48$\pm$5.53 &	63.48$\pm$5.53 &	63.48$\pm$5.53 &	\textbf{64.35$\pm$4.36} &	63.19$\pm$6.77 &	63.19$\pm$6.77 &	62.61$\pm$5.45 &	62.61$\pm$5.45 &	63.48$\pm$5.53 &	62.61$\pm$6.94 &	63.48$\pm$5.53 &	\textbf{64.35$\pm$4.36} \\
credit         &	74.18$\pm$0.35$\ddagger$ &	74.69$\pm$0.44 &	74.63$\pm$0.45 &	74.63$\pm$0.45 &	74.63$\pm$0.45 &	\textbf{74.72$\pm$0.35} &	\textbf{74.72$\pm$0.35} &	\textbf{74.72$\pm$0.35} &	74.70$\pm$0.44 &	74.69$\pm$0.44 &	74.63$\pm$0.45 &	\textbf{74.72$\pm$0.35} &	74.62$\pm$0.31 &	\textbf{74.72$\pm$0.35} \\
page           &	95.96$\pm$0.35 &	96.11$\pm$0.34 &	\textbf{96.11$\pm$0.29} &	\textbf{96.11$\pm$0.29} &	\textbf{96.11$\pm$0.29} &	\textbf{96.11$\pm$0.29} &	\textbf{96.11$\pm$0.29} &	\textbf{96.11$\pm$0.29} &	96.11$\pm$0.34 &	96.11$\pm$0.34 &	\textbf{96.11$\pm$0.29} &	\textbf{96.11$\pm$0.29} &	\textbf{96.11$\pm$0.29} &	\textbf{96.11$\pm$0.29} \\
shuttle        &	\textbf{99.84$\pm$0.03} &	99.83$\pm$0.02 &	99.83$\pm$0.03 &	99.83$\pm$0.03 &	99.83$\pm$0.03 &	99.83$\pm$0.03 &	99.83$\pm$0.03 &	99.83$\pm$0.03 &	99.83$\pm$0.02 &	99.83$\pm$0.02 &	99.83$\pm$0.03 &	99.83$\pm$0.03 &	99.83$\pm$0.03 &	99.83$\pm$0.03 \\
wilt           &	98.01$\pm$0.50 &	97.97$\pm$0.54 &	97.81$\pm$0.34 &	97.81$\pm$0.34 &	97.81$\pm$0.34 &	\textbf{98.04$\pm$0.38} &	97.75$\pm$0.34 &	97.75$\pm$0.34 &	97.97$\pm$0.54 &	97.97$\pm$0.54 &	97.81$\pm$0.34 &	97.91$\pm$0.48 &	97.95$\pm$0.47 &	\textbf{98.04$\pm$0.38} \\
ecoli          &	\textbf{96.36$\pm$1.55} &	95.76$\pm$1.13 &	95.76$\pm$1.13 &	95.76$\pm$1.13 &	95.76$\pm$1.13 &	95.76$\pm$1.13 &	95.76$\pm$1.13 &	95.76$\pm$1.13 &	95.76$\pm$1.13 &	95.76$\pm$1.13 &	95.76$\pm$1.13 &	95.76$\pm$1.13 &	95.76$\pm$1.13 &	95.76$\pm$1.13 \\
mammo\_graphic &	77.71$\pm$2.70 &	78.02$\pm$2.76 &	77.92$\pm$2.90 &	77.92$\pm$2.90 &	77.92$\pm$2.90 &	77.92$\pm$2.90 &	77.92$\pm$2.90 &	77.92$\pm$2.90 &	\textbf{78.12$\pm$2.81} &	78.02$\pm$2.76 &	77.92$\pm$2.90 &	77.92$\pm$2.90 &	77.71$\pm$3.06 &	77.92$\pm$2.90 \\
madelon        &	70.38$\pm$2.21 &	\textbf{70.46$\pm$1.96} &	69.69$\pm$2.05 &	69.69$\pm$2.05 &	69.69$\pm$2.05 &	70.31$\pm$1.89 &	69.92$\pm$1.58 &	69.92$\pm$1.58 &	70.15$\pm$1.96 &	\textbf{70.46$\pm$1.96} &	69.69$\pm$2.05 &	70.31$\pm$1.89 &	70.35$\pm$1.94 &	70.31$\pm$1.89 \\
sensor (4d)    &	\textbf{96.99$\pm$0.50} &	96.83$\pm$0.45 &	96.98$\pm$0.43 &	96.98$\pm$0.43 &	96.98$\pm$0.43 &	96.98$\pm$0.43 &	96.98$\pm$0.43 &	96.98$\pm$0.43 &	96.87$\pm$0.43 &	96.83$\pm$0.45 &	96.98$\pm$0.43 &	96.98$\pm$0.43 &	96.98$\pm$0.43 &	96.98$\pm$0.43 \\
sensor (24d)   &	\textbf{88.82$\pm$0.78} &	88.78$\pm$0.74 &	88.43$\pm$0.66 &	88.43$\pm$0.66 &	88.43$\pm$0.66 &	88.56$\pm$0.57 &	88.56$\pm$0.57 &	88.56$\pm$0.57 &	88.74$\pm$0.75 &	88.78$\pm$0.74 &	88.43$\pm$0.66 &	88.56$\pm$0.57 &	88.56$\pm$0.57 &	88.56$\pm$0.57 \\
waveform       &	85.03$\pm$0.83 &	85.09$\pm$0.60 &	85.07$\pm$0.78 &	85.07$\pm$0.78 &	85.07$\pm$0.78 &	\textbf{85.19$\pm$0.53} &	85.15$\pm$0.57 &	85.15$\pm$0.57 &	85.05$\pm$0.63 &	85.09$\pm$0.60 &	85.07$\pm$0.78 &	85.07$\pm$0.32 &	85.17$\pm$0.26 &	\textbf{85.19$\pm$0.53} \\
EEGEyeState    &	\textbf{96.72$\pm$0.20}$\dagger$ &	96.37$\pm$0.21 &	96.24$\pm$0.24 &	96.24$\pm$0.24 &	96.24$\pm$0.24 &	96.38$\pm$0.15 &	96.38$\pm$0.15 &	96.38$\pm$0.15 &	96.37$\pm$0.21 &	96.37$\pm$0.21 &	96.24$\pm$0.24 &	96.38$\pm$0.15 &	96.38$\pm$0.15 &	96.38$\pm$0.15 \\
	\hline
	$t$-test (W/T/L) & 1/12/1 & 0/14/0 & 0/14/0 & 0/14/0 & 0/14/0 & 0/14/0 & 0/14/0 & 0/14/0 & 0/14/0 & 0/14/0 & 0/14/0 & 0/14/0 & 0/14/0	& --- \\
	Average Rank & 6.64 & 7.68 & 9.18 & 9.18 & 9.18 & 5.00 & 6.89 & 6.89 & 8.50 & 7.68 & 9.18 & 7.18 & 6.82 & 5.00 \\
	\bottomrule
\end{tabular}
\end{threeparttable}
}
\end{table*}

\begin{table*}[th]
\centering
\caption{
Comparison of different diversity measures with \fdabbr{} using AdaBoost to produce an ensemble with SVMs as individual classifiers. 
The column named ``Ensem'' is the corresponding result for the entire ensemble without pruning. 
The best accuracy with a lower standard deviation is indicated with bold fonts for each data set (row).  
}\label{tab,frame:adaboost,SVM}
\renewcommand\tabcolsep{1.8pt}
\scalebox{0.641}{
\begin{threeparttable}[b]
\begin{tabular}{lc ccccc ccccccc c}
	\toprule
	Data Set & Ensem & Disag & QStat & Corre & KStat & DoubF & KWVar & Inter & EntCC & EntSK & Diffi & GeneD & CFail & \pdabbr{} \\ 
	\midrule
gmm (10d)   &	\textbf{99.80$\pm$0.25} &	99.80$\pm$0.40 &	99.80$\pm$0.40 &	99.80$\pm$0.40 &	99.80$\pm$0.40 &	99.80$\pm$0.40 &	99.80$\pm$0.40 &	99.80$\pm$0.40 &	99.80$\pm$0.40 &	99.80$\pm$0.40 &	99.80$\pm$0.40 &	99.80$\pm$0.40 &	99.80$\pm$0.40 &	99.80$\pm$0.40 \\
sonar       &	62.93$\pm$3.90 &	64.39$\pm$4.25 &	64.39$\pm$7.96 &	64.39$\pm$7.96 &	64.39$\pm$7.96 &	62.44$\pm$4.25 &	65.85$\pm$7.40 &	65.85$\pm$7.40 &	61.46$\pm$4.47 &	61.95$\pm$5.69 &	64.39$\pm$7.96 &	65.85$\pm$7.40 &	65.85$\pm$7.40 &	\textbf{66.83$\pm$7.00} \\
waveform    &	88.21$\pm$0.50$\ddagger$ &	89.97$\pm$0.52 &	89.81$\pm$0.48$\ddagger$ &	89.81$\pm$0.48$\ddagger$ &	89.81$\pm$0.48$\ddagger$ &	89.51$\pm$0.77$\ddagger$ &	89.51$\pm$0.77$\ddagger$ &	89.51$\pm$0.77$\ddagger$ &	89.97$\pm$0.52 &	89.97$\pm$0.52 &	89.81$\pm$0.48$\ddagger$ &	89.51$\pm$0.77$\ddagger$ &	89.51$\pm$0.77$\ddagger$ &	\textbf{90.71$\pm$0.75} \\
landsat     &	\textbf{65.24$\pm$0.00} &	\textbf{65.24$\pm$0.00} &	\textbf{65.24$\pm$0.00} &	\textbf{65.24$\pm$0.00} &	\textbf{65.24$\pm$0.00} &	\textbf{65.24$\pm$0.00} &	\textbf{65.24$\pm$0.00} &	\textbf{65.24$\pm$0.00} &	\textbf{65.24$\pm$0.00} &	\textbf{65.24$\pm$0.00} &	\textbf{65.24$\pm$0.00} &	\textbf{65.24$\pm$0.00} &	\textbf{65.24$\pm$0.00} &	\textbf{65.24$\pm$0.00} \\
page        &	90.59$\pm$0.21 &	90.99$\pm$0.51 &	90.95$\pm$0.48 &	90.95$\pm$0.48 &	90.95$\pm$0.48 &	90.95$\pm$0.48 &	90.95$\pm$0.48 &	90.95$\pm$0.48 &	90.86$\pm$0.49 &	90.86$\pm$0.49 &	90.95$\pm$0.48 &	90.95$\pm$0.48 &	90.97$\pm$0.44 &	\textbf{90.99$\pm$0.41} \\
shuttle     &	97.12$\pm$0.15 &	97.28$\pm$0.18 &	\textbf{97.28$\pm$0.16} &	\textbf{97.28$\pm$0.16} &	\textbf{97.28$\pm$0.16} &	\textbf{97.28$\pm$0.16} &	\textbf{97.28$\pm$0.16} &	\textbf{97.28$\pm$0.16} &	97.28$\pm$0.18 &	97.28$\pm$0.18 &	\textbf{97.28$\pm$0.16} &	\textbf{97.28$\pm$0.16} &	\textbf{97.28$\pm$0.16} &	95.25$\pm$2.08 \\
yeast       &	65.00$\pm$1.14 &	55.41$\pm$12.42 &	55.41$\pm$12.42 &	55.41$\pm$12.42 &	55.41$\pm$12.42 &	54.93$\pm$12.09 &	54.93$\pm$12.09 &	54.93$\pm$12.09 &	59.80$\pm$9.89 &	59.93$\pm$9.93 &	55.41$\pm$12.42 &	54.93$\pm$12.09 &	45.20$\pm$10.17$\ddagger$ &	\textbf{65.07$\pm$1.18} \\
sensor (4d) &	\textbf{85.63$\pm$0.90} &	84.47$\pm$0.78 &	84.00$\pm$1.45 &	84.00$\pm$1.45 &	84.00$\pm$1.45 &	77.58$\pm$8.33 &	84.01$\pm$0.85 &	84.01$\pm$0.85 &	84.47$\pm$0.78 &	84.47$\pm$0.78 &	84.00$\pm$1.45 &	81.21$\pm$7.51 &	81.28$\pm$7.54 &	84.01$\pm$0.85 \\
EEGEyeState &	\textbf{55.13$\pm$0.00} &	\textbf{55.13$\pm$0.00} &	\textbf{55.13$\pm$0.00} &	\textbf{55.13$\pm$0.00} &	\textbf{55.13$\pm$0.00} &	\textbf{55.13$\pm$0.00} &	\textbf{55.13$\pm$0.00} &	\textbf{55.13$\pm$0.00} &	\textbf{55.13$\pm$0.00} &	\textbf{55.13$\pm$0.00} &	\textbf{55.13$\pm$0.00} &	\textbf{55.13$\pm$0.00} &	\textbf{55.13$\pm$0.00} &	\textbf{55.13$\pm$0.00} \\
	\hline
	$t$-test (W/T/L) & 1/8/0 & 0/9/0 & 1/8/0 & 1/8/0 & 1/8/0 & 1/8/0 & 1/8/0 & 1/8/0 & 0/9/0 & 0/9/0 & 1/8/0 & 1/8/0 & 2/7/0	& --- \\
	Average Rank & 9.33 & 5.39 & 7.56 & 7.56 & 7.56 & 9.56 & 7.72 & 7.72 & 6.72 & 6.50 & 7.56 & 8.50 & 8.17 & 5.17 \\
	\bottomrule
\end{tabular}
\end{threeparttable}
}
\end{table*}

\begin{figure*}[tbh]
\centering 
\begin{minipage}{\linewidth}
\centering
\subfloat[]{
    \includegraphics[scale=0.5461]{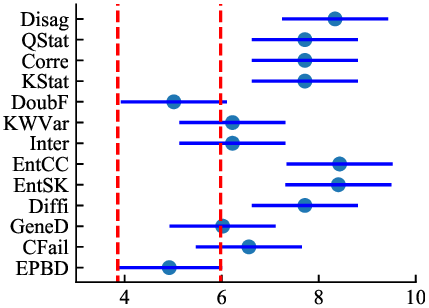}}
\hspace{0.5em}
\subfloat[]{
    \includegraphics[scale=0.5461]{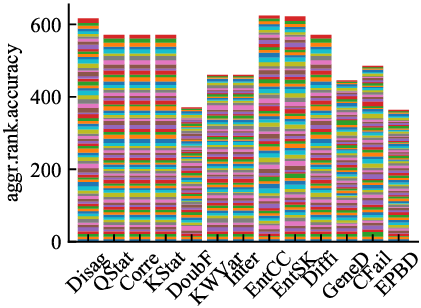}}
\hspace{0.5em}
\subfloat[]{\label{fig,expt3bi:bag,c}
    \includegraphics[scale=0.5461]{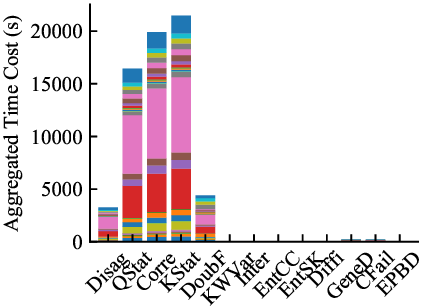}}
\hspace{0.5em}
\subfloat[]{
    \includegraphics[scale=0.5461]{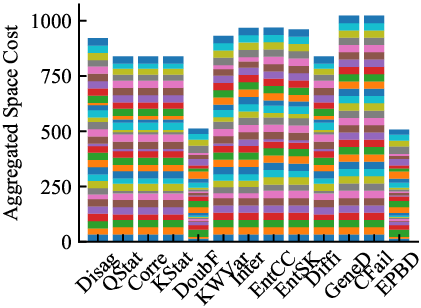}}
\caption{
Comparison of various diversity measures with \fdabbr{} and \pdabbr{} on the test accuracy, using bagging to produce homogeneous ensembles. 
(a) Friedman test chart (non-overlapping means significant difference) \citep{demvsar2006statistical}. 
(b) The aggregated rank for each diversity measure (the smaller the better) \citep{qian2015pareto}. 
(c) The aggregated time cost for each measure. 
(d) The aggregated space cost for each measure. 
}\label{fig,expt3:Bag,cost,NN}
\vspace{0.8em}
\end{minipage}
\begin{minipage}{\linewidth}
\centering
\subfloat[]{
    \includegraphics[scale=0.5461]{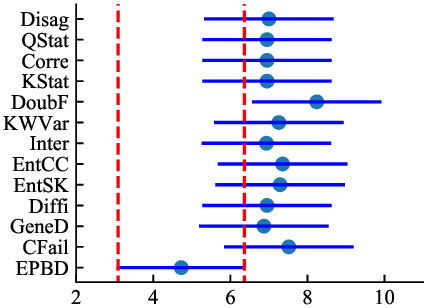}}
\hspace{0.5em}
\subfloat[]{
    \includegraphics[scale=0.5461]{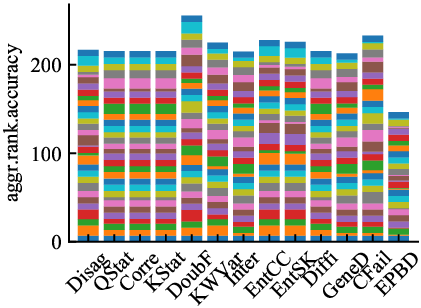}}
\hspace{0.5em}
\subfloat[]{\label{fig,expt3bi:ada,c}
    \includegraphics[scale=0.5461]{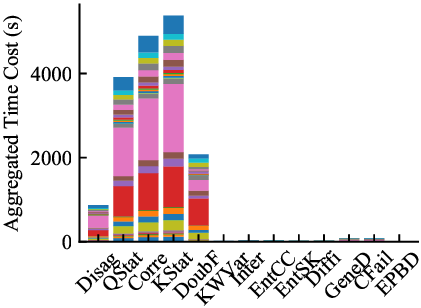}}
\hspace{0.5em}
\subfloat[]{
    \includegraphics[scale=0.5461]{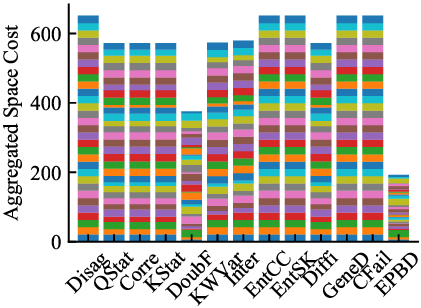}}
\caption{
Comparison of various diversity measures with \fdabbr{} and \pdabbr{} on the test accuracy, using AdaBoost to produce homogeneous ensembles. 
(a) Friedman test chart (non-overlapping means significant difference) \citep{demvsar2006statistical}. 
(b) The aggregated rank for each diversity measure (the smaller the better) \citep{qian2015pareto}. 
(c) The aggregated time cost for each measure. 
(d) The aggregated space cost for each measure. 
}\label{fig,expt3:Ada,cost,KNNu}
\end{minipage}
\end{figure*}

\subsection{Validating the Error Decomposition for Classification Ensembles}

In this subsection, experiments are conducted to verify the proposed error decomposition for classification ensembles (Theorem~\ref{th1}). 
To verify the effectiveness of the proposed diversity measure in Eq.~\eqref{eq:6,th1}, we compare the proposed diversity measure with other diversity measures in the literature, including %
the disagreement measure (Disag)~\citep{skalak1996sources,ho1998random},  
$Q$-statistic (QStat)~\citep{yule1900vii}, 
correlation coefficient (Corre)~\citep{sneath1973numerical},  
$\kappa$-statistic (KStat)~\citep{cohen1960coefficient},  
double-fault (DoubF)~\citep{giacinto2001design}, 
Kohavi-Wolpert variance (KWVar)~\citep{kohavi1996bias}, 
the interrater agreement (Inter)~\citep{fleiss1981statistical}, 
the entropy of the votes (EntCC~\citep{cunningham2000diversity} and EntSK~\citep{shipp2002relationships}), 
the difficulty index (Diffi)~\citep{hansen1990neural,kuncheva2003diversity}, 
the generalized diversity (GeneD)~\citep{partridge1997software}, and 
the coincident failure diversity (CFail)~\citep{partridge1997software}.
As shown in Figure~\ref{fig:decom:Diver,NN}\subref{fig:decom:Diver,NN,Bag} and Figure~\ref{fig:decom:Bag,NN}, the degree of correlation\footnote{%
The correlation coefficient used in this paper refers to the Pearson product-moment correlation coefficient \cite{wiki_corrcoef}, measuring the linear correlation between two variables $\mathsf{X}$ and $\mathsf{Y}$. 
Its value ranges from $-1$ to $1$. 
A value of $1$ is total positive linear correlation, $0$ is no linear correlation, and $-1$ is total negative linear correlation \cite{wiki_pearson_coeff}.
} between the loss difference $-(\bar{G}-\bar{A})$ and the proposed diversity measure $\bar{D}$ in Eq.~\eqref{eq:6c,th1} is higher than that between the loss difference and other diversity measures. 
Similar results are presented in Figure~\ref{fig:decom:Diver,NN}\subref{fig:decom:Diver,NN,Ada} and Figure~\ref{fig:decom:Ada,NN}. 
Therefore, we could suppose the proposed error decomposition for classification ensembles is reasonable.

\begin{figure*}[t]
\centering
    \subfloat[]{\centering
        \includegraphics[scale=0.5861]{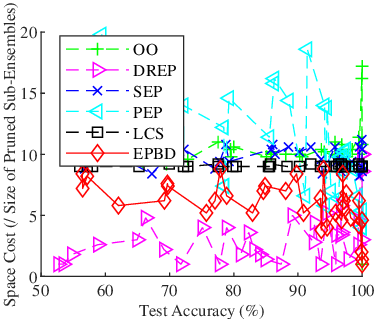}}
    \hspace{0.5em}
    \subfloat[]{\centering
        \includegraphics[scale=0.5861]{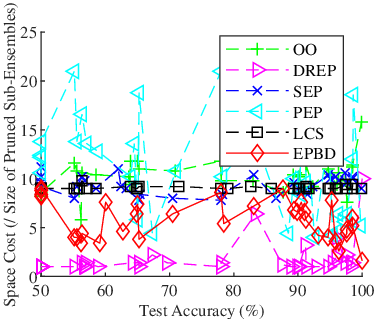}}
    \hspace{0.5em}
    \subfloat[]{\centering
        \includegraphics[scale=0.5861]{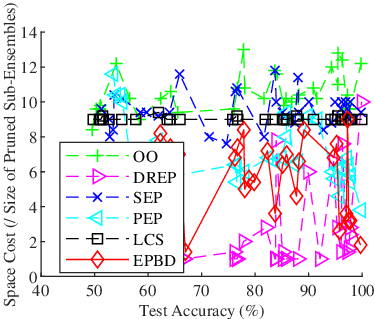}}
    \hspace{0.5em}
    \subfloat[]{\centering
        \includegraphics[scale=0.5861]{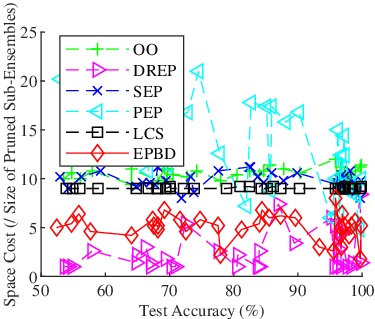}}
    \\
    \subfloat[]{\centering
        \includegraphics[scale=0.5861]{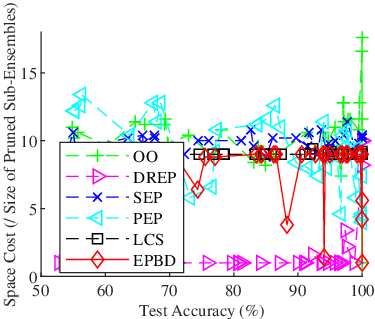}}
    \hspace{0.5em}
    \subfloat[]{\centering
        \includegraphics[scale=0.5861]{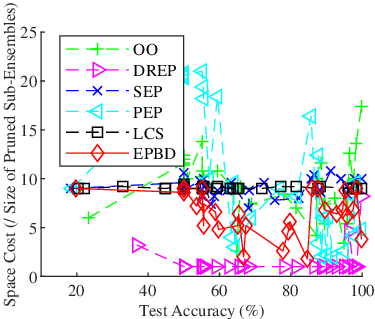}}
    \hspace{0.5em}
    \subfloat[]{\centering
        \includegraphics[scale=0.5861]{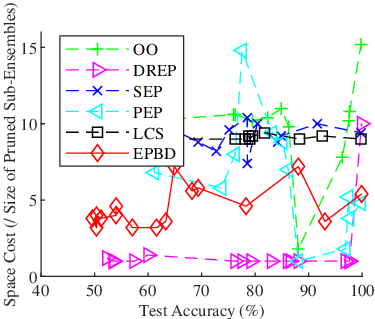}}
    \hspace{0.5em}
    \subfloat[]{\centering
        \includegraphics[scale=0.5861]{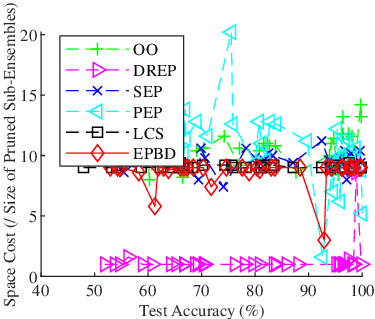}}
\caption{
Comparison of the space cost between \pdabbr{} and other methods that cannot fix the size of pruned sub-ensembles. 
Although \pdabbr{} cannot fix the number of individual classifiers in the pruned sub-ensemble, it could guarantee that the size of the pruned sub-ensemble is no more than the up limit of that, based on the pruning rate. 
(a--d) Using bagging with DTs, SVMs, LSVMs, and KNNs as individual classifiers, respectively. 
(e--h) Using AdaBoost with DTs, SVMs, LSVMs, and KNNs as individual classifiers, respectively. 
}\label{fig:space}
\end{figure*}

\subsection{Validating the Relationship Between the Proposed Diversity and Ensemble Performance}

In this subsection, experiments are conducted to verify the proposed relationship between the proposed diversity and generalization performance of classification ensembles (Theorem~\ref{th2}). 
Before that, we need to verify the ensemble performance's relationship with the estimated risk. 
The experimental results are reported in Figure~\ref{fig:eg1:B-partset}, including different individual classifiers. 
In each experiment, the estimated risk was calculated by Eq.~\eqref{eq:25}, based on the proposed diversity by Eq.~\eqref{eq:11}. 
The correlation between estimated risk and real error rate was calculated, as annotations shown in Figures~\ref{fig:eg1:B-partset}\subref{fig:eg1:B-NB-partset:a}--\ref{fig:eg1:B-partset}\subref{fig:eg1:B-LM2-partset:b}, presenting a high level of correlation between them. 
Besides, experimental results reported in Figures~\ref{fig:eg3:BagDT}--\ref{fig:eg3:AdaLM1} indicate that the relationship between the proposed diversity and ensemble performance in Theorem~\ref{th2} is faithful. 
Along with the increase of diversity, the ensemble performance would show an ``upward-downward-upward-downward'' trend in Figures~\ref{fig:eg3:BagDT}\subref{fig:eg3:BagDT:a} and \ref{fig:eg3:AdaLM1}\subref{fig:eg3:AdaLM1:a}, which coincides with the analyses in Section~\ref{propose:relation}. 
Therefore, we could utilize the proposed relationship between the proposed diversity and generalization error to guide our understanding of diversity, and this relationship is a good start to dig the real role of diversity in ensemble learning. 

\subsection{Comparison Between \pdabbr{} and the State-of-the-art Ensemble Pruning Methods}

In this subsection, we compare the performance of various ensemble pruning methods, including ES, KL, KP, OO, DREP, SEP, PEP, MRMR, MRMC, DiscEP, TSP.AP, TSP.DP, GMA, and LCS with the proposed \pdabbr{} method. 
Experimental results reported in Tables~\ref{tab,pru:bagging}--\ref{tab,pru:adaboost} contain the average test accuracy of each method and the corresponding standard deviation under $5$-fold cross-validation on each data set. 
For instance, each row (data set) in Table~\ref{tab,pru:bagging} compares the classification accuracy using bagging with the same type of individual classifiers, indicating results with higher accuracy and lower standard deviation by bold fonts. 
Besides, the significance of the difference in the accuracy performance between the two methods is examined by two-tailed paired $t$-tests at $5\%$ significance level to tell if two ensemble pruning methods have significantly different results. 
Two methods end in a tie if there is no significant statistical difference; otherwise, the one with higher values of accuracy will win. 
The performance of each method is reported in the last two rows of Table~\ref{tab,pru:bagging}, compared with \pdabbr{} in terms of the average rank and the number of data sets that \pdabbr{} has won, tied, or lost, respectively. 
We may notice that OO or PEP might achieve the best results instead of \pdabbr{} in a few cases. 
Yet \pdabbr{} could still reach an equal level of accuracy performance without significant difference ending up with a tie. 
Moreover, both of them cannot restrict the number of classifiers after ensemble pruning like \pdabbr{} does, leading to keeping more individual classifiers than we desire in most situations. 
Meanwhile, it could be inferred that \pdabbr{} could achieve competitive results even though it only keeps fewer individual classifiers, which means that the principle we used to guide the pruning process is effective and that utilizing our proposed relationship is reasonable. 
Figure~\ref{fig,expt4s:Bag,cost,NN}\subref{fig,expt4sbi:bag,a} shows that \pdabbr{} could achieve competitive results as OO, PEP, MRMR, and TSP.AP, and \pdabbr{} has significant superiority over other compared pruning methods. 
Figure~\ref{fig,expt4s:Bag,cost,NN}\subref{fig,expt4sbi:bag,b} presents the aggregated rank for each method, describing the similar conclusions to Figure~\ref{fig,expt4s:Bag,cost,NN}\subref{fig,expt4sbi:bag,a}. 
Figures~\ref{fig,expt4s:Bag,cost,NN}\subref{fig,expt4sbi:bag,c}--\ref{fig,expt4s:Bag,cost,NN}\subref{fig,expt4sbi:bag,d} illustrate the aggregated time cost and the aggregated space cost for each pruning method, respectively. 
It could be inferred that PEP, DiscEP, and LCS usually cost huge processing time to do ensemble pruning tasks. 
Similar results are reported in Figure~\ref{fig,expt4s:Ada,cost,KNNd}. 

\subsection{Comparison Between \pdabbr{} and \fdabbr{} Using Other Diversity Measures}

In this subsection, we compare the performance of \fdabbr{} with various diversity measures involved, including pairwise measures (Disag, QStat, Corre, KStat, and DoubF) and non-pairwise measures (KWVar, Inter, EntCC, EntSK, Diffi, GeneD, and CFail) with the proposed \pdabbr{} method. 
Experimental results reported in Tables~\ref{tab,frame:bagging,NB}--\ref{tab,frame:adaboost,SVM} contain the average test accuracy of each method and the corresponding standard deviation under $5$-fold cross-validation on each data set. 
For instance, each row (data set) in Table~\ref{tab,frame:bagging,NB} compares the classification accuracy using bagging with the same type of individual classifiers, indicating results with higher accuracy and lower standard deviation by bold fonts. 
Besides, the significance of the difference in the accuracy performance between two methods is examined by two-tailed paired $t$-tests at $5\%$ significance level to tell if two diversity measures used in \fdabbr{} have significantly different results. 
Two measures end in a tie if there is no significant statistical difference; otherwise, the one with higher values of accuracy will win. 
The performance of each measure is reported in the last two rows of Table~\ref{tab,frame:bagging,NB}, compared with \pdabbr{} using the proposed diversity measure in this paper in terms of the average rank and the number of data sets that \pdabbr{} has won, tied, or lost, respectively. 
It could be inferred that \pdabbr{} could achieve competitive results with \fdabbr{} using other diversity measures according to the results of two-tailed paired $t$-tests in Tables~\ref{tab,frame:bagging,NB}--\ref{tab,frame:adaboost,SVM}. 
The similar observations also appear in Figures~\ref{fig,expt3:Bag,cost,NN}--\ref{fig,expt3:Ada,cost,KNNu}. 
Moreover, Figures~\ref{fig,expt3:Bag,cost,NN}--\ref{fig,expt3:Ada,cost,KNNu} present the time cost and space cost for each diversity measure. 
It could be inferred that pairwise diversity measures used in \fdabbr{} usually cost much more time during the pruning process, as shown in Figure~\ref{fig,expt3:Bag,cost,NN}\subref{fig,expt3bi:bag,c} and Figure~\ref{fig,expt3:Ada,cost,KNNu}\subref{fig,expt3bi:ada,c}.

\subsection{Comparison of Space Cost Between \pdabbr{} and Methods that Cannot Fix the Size of Pruned Sub-Ensembles}

In this subsection, we compare the accuracy and the space cost between \pdabbr{} with methods that cannot fix the size of pruned sub-ensembles, such as OO, DREP, SEP, PEP, and LCS. 
Experimental results are reported in Figure~\ref{fig:space}. 
As shown in Figure~\ref{fig:space}, DREP only keeps one of the individual classifiers that is not an ensemble at all in most cases; 
OO, SEP, and PEP usually generate pruned sub-ensembles with a larger size than the up limit of that; 
the size of pruned sub-ensembles generated by LCS is relatively steady around the up limit of that.  
Besides, although \pdabbr{} cannot fix the number of individual classifiers in the pruned sub-ensemble, it could guarantee that the size of the pruned sub-ensemble is no more than the up limit of that, based on the pruning rate. 
More comparison of space cost could be referred to Figure~\ref{fig,expt4s:Bag,cost,NN}\subref{fig,expt4sbi:bag,d} and Figure~\ref{fig,expt4s:Ada,cost,KNNd}\subref{fig,expt4sbi:ada,d}. 
Therefore, we could conclude that \pdabbr{} could run as expected to generate a comparable ensemble performance with a smaller size to the original ensemble.

\section{Conclusion}
\label{conclusion}

This paper has investigated and utilized diversity in classification ensembles, and it made the following contributions to the ensemble learning community. 
First of all, inspired by the regression ensembles, this paper proposed the measure of diversity utilizing the error decomposition for classification ensembles, which broke the error of classification ensembles into two terms: the accuracy and the diversity terms. 
The empirical results have confirmed that ``$\diver$'' is a distinct diversity measure. 
Secondly, we have theoretically investigated the relationship between the proposed diversity and generalization of the ensemble through the bound between margin and generalization. 
These theoretical analyses solve an open question in the ensemble area to some extent, i.e., when does diversity help the generalization of ensembles? 
Based on the analyses, the relationship between the proposed diversity and generalization error varies when diversity is in different ranges. 
In some ranges, more diversity could lead to better generalization, while in other ranges, more diversity is not beneficial to the generalization based on the bound analyses. 
Thirdly, in order to validate and employ the relationship of diversity to improve the performance of ensembles, we have proposed \pdabbr{} that can prune an ensemble without much performance degradation. 
Besides, an ensemble pruning framework (\ie{} \fdabbr{}) is proposed to utilize the trade-off between accuracy and diversity and generate pruned sub-ensembles with a smaller size. 
Note that in \fdabbr{}, not only the proposed diversity measure in this paper but also other existing diversity measures could be employed. 
Although this generalization bound could be loose, our work provides a direction and one theoretical attempt to reveal the impacts of diversity on the generalization. 
Our future work is to generalize the proposed methodology to the multi-classification problems. 

\appendix[Example of Extended Analyses in Multi-Classification Problems]
\label{method:multi-class}

In this section, we give a short example to extend the aforementioned analyses in Section~\ref{methodology} to multi-classification problems, that means $\mathcal{Y}=\{0,1,...,n_c-1\}$ with $n_c \,(n_c\geqslant 2 \,,$\; $n_c\in\mathbb{Z}^+)$ different labels. 
Individual classifiers in an ensemble $(F=\{f_1,...,f_{|F|}\})$ have been trained and are combined by weighted voting, \ie{} 
\begin{equation}%
\small
    f_{ens}(\mathbf{x}) = \argmax_{c_k\in\mathcal{Y}} \sum_{i\in[|F|]} w_i\mathbb{I}\big( f_i(\mathbf{x})=c_k \big)
    \,,\label{eqa:1}
\end{equation}
where $w_i$ is the weight corresponding to the individual classifier $f_i$, satisfying that $\sum_{i\in[|F|]} w_i=1,\, \forall\, w_i\geqslant 0$\,. 
To extend the aforementioned analyses in Section~\ref{methodology} to multi-classification problems, we give the extended proposed error decomposition and diversity measure as a short example. 
To this end, we need to extend the \emph{margin} of an ensemble to multi-classification problems first. 
Inspired by \citep{schapire1998boosting}, we introduce a new concept named as ``\emph{credence}'' of one classifier $f(\cdot)$, represented by $\dist(\cdot)$, as mappings from $\mathcal{X}\times\mathcal{Y}$ to $[0,1]$, with the interpretation that the result $f(\mathbf{x})$ predicted by $f$ is more plausible for the label $y$ of the instance $\mathbf{x}$ if $\dist(f,\mathbf{x},y)$ is closer to $1$. 
Besides, the credence of the ensemble $f_{ens}$ on one instance $(\mathbf{x},y)$ is defined as 
\begin{equation}%
\small
    \dist(f_{ens},\mathbf{x},y) = \sum_{i\in[|F|]} w_i\cdot \dist(f_i,\mathbf{x},y)
    \,,\label{eqa:2}
\end{equation}
therefore the ensemble predicts label $y$ for the instance $\mathbf{x}$ if $\dist(f_{ens},\mathbf{x},y) > \max_{y'\neq y,\, y'\in\mathcal{Y}} \dist(f_{ens},\mathbf{x},y')$ (and ties are broken arbitrarily). 
Subsequently, the \emph{margin} of the ensemble on the instance is defined as 
\begin{equation}%
\small
    \margin(f_{ens},\mathbf{x}) = \dist(f_{ens},\mathbf{x},y) - \max_{y'\neq y,\, y'\in\mathcal{Y}} \dist(f_{ens},\mathbf{x},y')
    \,,\label{eqa:3}
\end{equation}
and that of one individual classifier $f_i$ could be obtained similarly. 
Consequently, $f_{ens}$ would give the wrong prediction on $(\mathbf{x},y)$ only if $\margin(f_{ens},\mathbf{x}) \leqslant 0$. 
Similar to Section~\ref{propose:measure}, we define the error function as 
\begin{equation}
\small
    \Err(f_{ens},\mathbf{x}) = \frac{1}{2}\big( 1-\margin(f_{ens},\mathbf{x}) \big)
    \,,\label{eqa:4}
\end{equation}
and the diversity measure would be 
\begin{equation}%
\small
    \diver(f_{ens},\mathbf{x}) = %
    \sum_{i\in[|F|]} w_i\cdot \Err(f_i,\mathbf{x}) - \Err(f_{ens},\mathbf{x})
    \,.\label{eqa:5}
\end{equation}%
The remaining issue is the definitions of \emph{credence}. 
As we know in binary classification problems where $\mathcal{Y}=\{-1,+1\}$, the \emph{credence} of one classifier $f$ is supposed to be viewed as %
\begin{equation}
\small
    \dist(f,\mathbf{x},y) = \frac{f(\mathbf{x})y+1}{2}
    \,.\label{eqa:6}
\end{equation}
As for multi-classification problems where $\mathcal{Y}=\{0,1,...,n_c-1\} \, (n_c\geqslant 2,\, n_c\in\mathbb{Z}^+)$, three functions could be proposed as possible forms of \emph{credence}, \ie{} %
\begin{subequations}
\begin{small}
\begin{align}
    \dist{}_1(f,\mathbf{x},y) =& \mathbb{I}\big( f(\mathbf{x}) =y \big) \,,\\
    \dist{}_2(f,\mathbf{x},y) =& 1-\frac{ |f(\mathbf{x})-y| }{n_c-1} \,,\\
    \dist{}_3(f,\mathbf{x},y) =& 1-\frac{ \big(f(\mathbf{x})-y\big)^2 }{(n_c-1)^2} \,.
\end{align}%
\end{small}%
\end{subequations}%
Consequently, the difference between the ensemble and its individual members, defined as the ``diversity'' measure, could be obtained by
\begin{scriptsize}
\begin{align}
    \diver(f_{ens},\mathbf{x}) 
    =& -\bigg( \Err(f_{ens},\mathbf{x}) - \sum_{i\in[|F|]} w_i\cdot \Err(f_i,\mathbf{x}) \bigg) \nonumber\\
    =& \sum_{i\in[|F|]} w_i\cdot \frac{ \margin(f_{ens},\mathbf{x}) - \margin(f_i,\mathbf{x}) }{2} \nonumber\\
    =& \sum_{i\in[|F|]} w_i\cdot \frac{ \dist(f_{ens},\mathbf{x},y) - \dist(f_i,\mathbf{x},y) }{2} \nonumber\\
    -& \sum_{i\in[|F|]} \frac{w_i}{2}\bigg( \max_{y'\neq y} \dist(f_{ens},\mathbf{x},y') - \max_{y'\neq y}\dist(f_i,\mathbf{x},y') \bigg) ,\label{eqa:10}%
\end{align}%
\end{scriptsize}%
that is, 
\begin{small}
\begin{align}
    \diver(f_{ens},\mathbf{x}) 
    =& \frac{1}{2} \sum_{i\in[|F|]} w_i\cdot \max_{y'\neq y}\dist(f_i,\mathbf{x},y') \nonumber\\
    -& \frac{1}{2} \max_{y'\neq y}\dist(f_{ens},\mathbf{x},y') 
    \,.\label{eqa:11}
\end{align}%
\end{small}%
Analogous results like Theorem~\ref{th1} could be obtained subsequently. 
Notice that the appendix is to point out one possibility of extending our theoretical results to multi-classification problems, which is a start. 
Further investigations and careful experiments are still needed before getting a more promising conclusion.

%

\ifCLASSOPTIONcaptionsoff
  \newpage
\fi

\bibliographystyle{IEEEtran}
\bibliography{title_abbr,refs}
\end{document}